\documentclass{article}
\usepackage{iclr2026_conference,times}

\usepackage{amsmath,amssymb}
\usepackage{booktabs}
\usepackage{graphicx}
\usepackage{placeins}
\usepackage{float}
\usepackage{microtype}
\usepackage{xcolor}
\usepackage{hyperref}
\usepackage{url}
\usepackage{fancyvrb}

\usepackage{amsmath,amsfonts,bm}

\def\eqref#1{equation~\ref{#1}}

\def\1{\bm{1}}

\DeclareMathAlphabet{\mathsfit}{\encodingdefault}{\sfdefault}{m}{sl}
\SetMathAlphabet{\mathsfit}{bold}{\encodingdefault}{\sfdefault}{bx}{n}

\newcommand{\softmax}{\mathrm{softmax}}

\newcommand{\method}{CALIPER}

\title{\method: Evidence-Gated LLM Prior Layers for Multi-Objective Bayesian Optimization}

\author{
  Jiangyu Chen\thanks{Both authors contributed equally to this research.} \quad Yi Ban\footnotemark[1] \\
  State Key Laboratory for Novel Software Technology, Nanjing University \\
  \texttt{jianyuchen@smail.nju.edu.cn} \\
  \And
  Tianfan Fu\thanks{Corresponding author.} \\
  State Key Laboratory for Novel Software Technology, Nanjing University \\
  \texttt{futianfan@nju.edu.cn} 
}

\newcommand{\hv}{\mathrm{HV}}
\newcommand{\qwenflash}{\mbox{Qwen3.6-Flash}}
\newcommand{\qwenmax}{\mbox{Qwen3.7-Max}}
\newcommand{\deepseek}{\mbox{DeepSeek-V4-Pro}}
\newcommand{\obsset}{\mathcal{D}}
\newcommand{\obsx}{\mathcal{X}^{\mathrm{obs}}}
\DeclareMathOperator{\clip}{clip}
\DeclareMathOperator{\corr}{corr}

\begin{document}
\maketitle
\begin{abstract}
Large language models (LLMs) are increasingly used as heuristic advisors for black-box optimization, yet their suggestions and self-reported confidence are not calibrated observations. This mismatch is especially visible in multi-objective Bayesian optimization (MOBO), where an LLM expert may be useful for one objective, misleading for another, and overconfident in both.

We propose \method{} (\emph{Counterfactual Adaptive LLM-Integrated Prior Evidence Replay}), an evidence-gated mechanism for turning LLM expert outputs into objective-wise Bayesian priors for discrete MOBO. Instead of treating the LLM as an optimizer or as a fixed acquisition heuristic, \method{} turns multiple LLM expert roles into residual prior mean functions. A reputation market learns expert-objective trust online from observed objective feedback, while decoupled counterfactual gates decide whether to use the prior without confidence, use it with confidence, or abstain from the prior entirely. The resulting prior layer shifts residual Gaussian-process posteriors back to the original objective space and can be used with standard MOBO acquisitions such as qLogNEHVI and qLogEHVI.

Across controlled synthetic stress tests, candidate-pool scaling diagnostics, and molecule optimization benchmarks with cached \qwenflash{} priors, we find that calibrated LLM priors improve robustness over fixed priors and can improve qLogNEHVI on large synthetic candidate pools. In a real molecule same-acquisition comparison, qLogNEHVI+\method{} improves or matches qLogNEHVI on ESOL, FreeSolv, and Lipophilicity, while fixed priors can help or hurt depending on the dataset. The real molecule results also show the main boundary condition: gains depend strongly on prior quality. On Lipophilicity, raw LLM priors are systematically biased, while a mixed expert committee gives a small calibrated gain and an oracle-style controlled prior shows that the layer can use high-quality prior information. These results support using LLMs in BO as checked Bayesian prior sources rather than as fixed suggestion engines.
\end{abstract}

\section{Introduction}

Bayesian optimization (BO) is a sample-efficient framework for optimizing expensive black-box functions, where each evaluation may require a laboratory experiment, a simulation, or a costly model call \citep{jones1998ego,frazier2018tutorial}. BO works by fitting a probabilistic surrogate model and using an acquisition function to trade off exploitation and exploration \citep{frazier2018tutorial}. Large language models (LLMs) provide broad prior knowledge over natural-language and structured inputs, and recent work has begun to use them in optimization loops \citep{liu2024llambo,chen2023instructzero}.

Recent LLM--BO work has explored this opportunity in several directions. \citet{liu2024llambo} frame BO histories in natural language and use an LLM to propose and evaluate candidates. \citet{bopro2025} use BO to adapt the search strategy of in-context LLM optimization. \citet{funbo2025} use an LLM-based program search procedure to discover acquisition functions. Other work uses BO as an outer-loop tool for tuning LLM systems, such as prompt selection, checkpoint merging, model fusion, compression, or data-mixture design \citep{hbbops2025,checkpoint2025,modelfusion2024,adaptivecompression2024,datamixture2025}. These directions leave a complementary question underexplored: if an LLM is not the optimizer itself, can it still serve as a useful Bayesian prior inside BO?

Treating LLM output as a prior is attractive because even an imperfect prior can accelerate search in the low-data regime. It is also risky. LLM scores are not calibrated observations, and self-reported confidence should not be assumed to match empirical correctness \citep{guo2017calibration,lin2022uncertaintywords,kadavath2022know}. An expert role can be useful for one objective but misleading for another; several roles can make correlated mistakes; and confidence may express linguistic certainty rather than statistical reliability. In multi-objective BO, these failure modes are especially visible: a single global trust weight can hide objective-specific expertise, while a fixed confidence rule can help one objective and hurt another.

This paper studies LLMs as uncertain, falsifiable, objective-wise Bayesian priors. Our premise is simple: an LLM prior should influence BO only to the extent that it earns trust from observed objective feedback. It should be updated separately for each expert and objective, its confidence should be treated as another uncertain signal, and the optimizer should be able to attenuate or abstain from the prior when evidence contradicts it. Crucially, this calibration mechanism should not require replacing the MOBO acquisition function; it should be a prior layer that can attach to strong acquisitions.

We instantiate this idea in \method{} (\emph{Counterfactual Adaptive LLM-Integrated Prior Evidence Replay}). Multiple LLM expert roles produce objective-wise prior scores and self-reported confidence values for every candidate. The BO loop fits residual Gaussian-process surrogates after subtracting an aggregated LLM prior. Expert-objective trust is updated online through a reputation market, and a decoupled counterfactual gate decides whether to use the prior without confidence, use it with confidence, or drop the prior entirely. Because the expert market evolves deterministically, the gate can replay counterfactual confidence decisions after each observation instead of waiting for high-variance long-horizon rewards. We implement the calibrated prior as a posterior-shift adapter, allowing the same layer to plug into qLogNEHVI and qLogEHVI.

\paragraph{Contributions.}
Our contributions are:
\begin{enumerate}
  \item We formulate LLM-generated expert predictions as residual priors for multi-objective BO, with trust indexed by both expert and objective.
  \item We introduce \method{}, an evidence-gated prior layer combining an objective-wise reputation market, deterministic counterfactual replay, and explicit prior abstention.
  \item We show how the calibrated prior layer can wrap residual GP posteriors and plug into strong qLogNEHVI and qLogEHVI acquisitions, rather than replacing them with a custom acquisition rule.
  \item We provide synthetic scaling, molecule, and prior-quality diagnostics showing both the promise and the boundary condition: calibrated priors can improve strong MOBO baselines when the prior contains useful signal, while fixed or biased LLM priors can hurt.
\end{enumerate}

\section{Related Work}

\paragraph{Bayesian optimization and MOBO.}
Bayesian optimization uses a surrogate model and an acquisition function to optimize expensive black-box objectives \citep{jones1998ego,frazier2018tutorial}. Multi-objective BO extends this setting to competing objectives; common baselines include scalarization methods such as ParEGO \citep{knowles2006parego} and hypervolume-based acquisitions such as qEHVI and qNEHVI \citep{daulton2020qehvi,daulton2021qnehvi}. We use these acquisitions through BoTorch \citep{balandat2020botorch}. Our method does not propose a new acquisition. It changes how external prior information enters the surrogate.

\paragraph{LLMs in optimization loops.}
Recent top-conference work has assigned several different roles to LLMs in BO. \citet{liu2024llambo} use LLMs for candidate proposal and evaluation inside model-based BO. \citet{bopro2025} use the probabilistic view of BO to adapt in-context LLM search. \citet{funbo2025} ask LLMs to discover acquisition functions written as code, while \citet{cake2025} use LLMs for kernel design. A related but different line uses BO to tune LLM systems, including prompt selection, checkpoint merging, model fusion, compression, and data-mixture design \citep{chen2023instructzero,hbbops2025,checkpoint2025,modelfusion2024,adaptivecompression2024,datamixture2025}. Our setting gives the LLM a different role: it does not choose the next candidate and does not replace the acquisition. It supplies prior scores that are checked against objective observations during BO.

\paragraph{Confidence and expert weighting.}
Confidence calibration asks whether predicted confidence matches empirical correctness \citep{guo2017calibration}. LLMs make this harder because confidence may be reported in words or prompted scores rather than model probabilities \citep{lin2022uncertaintywords,kadavath2022know}. Our reputation updates are also related to online learning with expert advice \citep{freund1997decision,cesabianchi2006prediction}. The difference is that feedback arrives only at BO-selected candidates, and each expert can be reliable for one objective but not another. This motivates expert-objective weights and counterfactual gates rather than a single global LLM confidence rule.

\section{Method}
\label{sec:method}

\subsection{Problem Formulation}

We consider discrete multi-objective black-box optimization. Let the candidate set be
\[
  \mathcal{X} = \{x_1,\ldots,x_N\}.
\]
Here \(\mathcal{X}\) is a finite candidate pool, \(x_i\) denotes the \(i\)-th candidate, and \(N\) is the pool size. Each candidate has an unknown objective vector
\[
  \mathbf{f}(x) = (f_1(x), \ldots, f_m(x)).
\]
Here \(m\) is the number of objectives and \(f_j(x)\) is the value of objective \(j\) at candidate \(x\). The optimizer begins with a small initial design and sequentially selects candidates to evaluate. The goal is to maximize multi-objective performance under a limited evaluation budget. In our experiments, \(m=2\), and we measure performance by final hypervolume, area under the hypervolume curve (AUC hypervolume), and best objective-sum.

We assume access to a set of LLM/expert priors. For each candidate \(x\), expert \(e\), and objective \(j\), the expert provider returns a predicted normalized objective score \(\mu_{ej}(x)\) and a scalar self-reported confidence value \(c_{ej}(x)\). In the cached priors used here, objective scores are normalized to \([0,1]\), and confidence values are parsed as scalar values in \([0,1]\). The symbol \(\mu_{ej}(x)\) is not an observation; it is an LLM-generated prior score. The value \(c_{ej}(x)\) is not treated as a calibrated probability. The optimizer must decide which experts to trust, for which objectives, and whether self-reported confidence should influence prior aggregation or online expert-weight updates.

\subsection{LLM Experts as Objective-Wise Priors}

We instantiate multiple expert roles, such as solubility-oriented, drug-likeness-oriented, or balanced absorption, distribution, metabolism, excretion, and toxicity (ADMET) experts. Each expert returns objective-wise scores and confidence values. These predictions are cached and used as prior information during BO.

Given trust weights \(\alpha_{ej,t}\), where \(\alpha_{ej,t}\) denotes the current reliability weight assigned to expert \(e\) for objective \(j\) at BO step \(t\), the aggregated prior for objective \(j\) is
\[
  p_{j,t}(x) =
  \frac{\sum_e \alpha_{ej,t}\,\tilde{c}_{ej,t}(x)\,\mu_{ej}(x)}
       {\sum_e \alpha_{ej,t}\,\tilde{c}_{ej,t}(x)},
\]
where \(p_{j,t}(x)\) is the LLM-derived prior mean for objective \(j\) at step \(t\), and \(\tilde{c}_{ej,t}(x)\) is the effective confidence after optional calibration or gating. The surrogate model is fit to residuals:
\[
  f_j(x) = p_{j,t}(x) + r_j(x).
\]
Here \(r_j(x)\) is the residual function that remains after subtracting the LLM prior from the true objective. This lets the optimizer use LLM prior structure while still correcting it with observed objective values.

\subsection{Objective-Wise Reputation Market}

The main strategy maintains a capital value for each expert-objective pair:
\[
  K_{ej}.
\]
Here \(K_{ej}\) is a scalar reputation account; higher capital means that expert \(e\) has recently predicted objective \(j\) more accurately.
At an observed point \((x_t,\mathbf{y}_t)\), we compute standardized error
\[
  \epsilon_{ej,t} = \frac{|\mu_{ej}(x_t)-y_{j,t}|}{s_{j,t}},
\]
where \(s_{j,t}\) is a running objective scale. A soft success score is
\[
  q_{ej,t} = \exp(-\tfrac{1}{2}\epsilon_{ej,t}^2).
\]
Thus \(q_{ej,t}\) is close to one for accurate predictions and close to zero for large scaled errors.
The capital update uses a proper-score-like reward:
\[
  R_{ej,t} = \clip(0.5 - 0.5\epsilon_{ej,t}^2, -2.0, 0.5),
\]
\[
  K_{ej,t+1} = (1-\lambda)K_{ej,t} + \eta\,\tilde{c}_{ej}(x_t)\,R_{ej,t}.
\]
Here \(\lambda\) is the capital discount factor, \(\eta\) is the reward step size, and \(\tilde{c}_{ej}(x_t)\) is the effective confidence multiplier used for this update. Capital is converted into relative expert weights through a temperature-controlled softmax across experts. A market-level trust gate shrinks all prior weights when the expert market has low absolute reputation.

\subsection{Residual Prior Layer}

Given the current prior \(p_{j,t}(x)\), we fit a Gaussian-process residual surrogate to
\[
  r_{j,\tau}=y_{j,\tau}-p_{j,t}(x_\tau).
\]
Here \(p_{j,t}\) is the prior available at BO step \(t\), and \(r_{j,\tau}\) is the residual target for the previously evaluated candidate \(x_\tau\).
The posterior prediction for objective \(j\) is reconstructed as
\[
  \hat{f}_{j,t}(x)=p_{j,t}(x)+m^r_{j,t}(x),
\]
where \(m^r_{j,t}\) is the residual posterior mean. More generally, if the residual model gives a posterior
\[
  r_j(x)\mid\obsset_t \sim \mathcal{N}(m^r_{j,t}(x), v^r_{j,t}(x)),
\]
then the prior-shifted posterior exposed to the acquisition function is
\[
  f_j(x)\mid\obsset_t \sim \mathcal{N}(p_{j,t}(x)+m^r_{j,t}(x), v^r_{j,t}(x)).
\]
Thus the LLM prior shifts the posterior mean but does not artificially shrink posterior uncertainty. If the prior is useful, it moves the acquisition toward promising regions; if it is harmful, the prior gate can shrink or drop \(p_{j,t}\), and the residual model falls back toward ordinary GP-based MOBO.

\subsection{Acquisition-Agnostic Posterior Shift}

\method{} is implemented as a prior layer rather than as a replacement acquisition function. The layer owns the expert-prior aggregation and returns \(p_{j,t}(x)\). A wrapped residual model then exposes the shifted posterior above to any acquisition that consumes a standard multi-output posterior. In the implementation, this is the role of the \texttt{PriorShiftedModel} adapter.

For strong MOBO baselines, we fit the residual GP to \(\{(x_\tau,y_\tau-p_t(x_\tau))\}\) and pass the shifted posterior to qLogEHVI or qLogNEHVI, the BoTorch log-improvement variants of qEHVI and qNEHVI \citep{daulton2020qehvi,daulton2021qnehvi,balandat2020botorch}. The acquisition therefore optimizes expected hypervolume improvement in the original objective space, while all LLM influence is confined to the calibrated prior mean. This design separates two questions that are often conflated: how to compute a good MOBO acquisition, and how much LLM prior information should be trusted.

We also use a randomized scalarized UCB rule as a lightweight diagnostic acquisition for residual-prior variants:
\[
  A_t(x)=\sum_{j=1}^m w_{j,t}\left(\hat{f}_{j,t}(x)+\beta_t s^r_{j,t}(x)\right)
  +\gamma(1-\bar{\alpha}_t)\sum_{j=1}^m w_{j,t}D_j(x).
\]
Here \(w_t\) is a random objective weight vector, \(s^r_{j,t}\) is residual posterior uncertainty, \(D_j(x)\) is expert disagreement for objective \(j\), \(\bar{\alpha}_t\) is average market trust, \(\beta_t=1+1.25(1-\bar{\alpha}_t)\), and \(\gamma=0.30\). When expert trust is low, this diagnostic acquisition explores more and gives additional value to candidates where experts disagree. The qLogNEHVI prior-layer results do not depend on this scalarized rule.

\subsection{Confidence Roles and Diagnostic Variants}

We do not assume that LLM confidence is calibrated. Before fixing the final gate, we evaluate raw confidence, no confidence, softened confidence, prior-only confidence, update-only confidence, and adaptive confidence gates as diagnostics. These variants are not separate proposed methods; they identify where confidence is useful and motivate the final decoupled gate.

For one adaptive baseline, the optimizer tracks whether confidence predicts expert success. For each objective, it maintains an online pooled correlation between reported confidence and soft success:
\[
  \rho_j = \corr(c_{ej,t}, q_{ej,t}).
\]
The gate is
\[
  g_j = \sigma(a_\rho(\rho_j-\tau_\rho)).
\]
When the correlation is non-positive or insufficiently supported, the gate closes. Effective confidence is
\[
  \tilde{c}_{ej}(x) = 1 + g_j(c_{ej}(x)-1).
\]
Thus \(g_j=0\) ignores confidence and \(g_j=1\) uses raw confidence.

Confidence can influence two parts of the algorithm:
\begin{enumerate}
  \item prior aggregation, which changes the prior score used to rank candidates;
  \item reputation updates, which changes how experts are rewarded or punished.
\end{enumerate}
The final \method{} gate therefore separates these roles instead of applying a single confidence multiplier everywhere.

\subsection{Decoupled Counterfactual Gate}

The \method{} gate keeps the reputation-market update but replaces a single confidence gate with two counterfactual decisions. This is the key difference from earlier confidence-phase variants: it does not merely decide whether to multiply by self-reported confidence. It asks two separate evidence questions: whether the LLM prior should enter the surrogate at all, and whether confidence should scale future expert-credit updates. The prior gate controls how the aggregated LLM prior enters the residual surrogate. For each objective \(j\), it compares three prior actions:
\[
  a \in \{\mathrm{no\_conf},\mathrm{conf},\mathrm{drop}\}.
\]
The first action aggregates expert priors with confidence set to one, the second uses reported confidence, and the third sets the prior contribution to zero. For action \(a\), define the historical residuals
\[
  r^{(a)}_{j,\tau}=y_{j,\tau}-p^{(a)}_{j,t}(x_\tau),
\]
and score them with a fixed-kernel GP marginal likelihood
\[
  S_{j,t}(a)=\frac{1}{t}\log p(r^{(a)}_{j,1:t}\mid \obsx_t).
\]
The prior-arm logits are relative to the conservative no-confidence arm:
\[
  \ell_{j,t}(a)=\eta_p\left(S_{j,t}(a)-S_{j,t}(\mathrm{no\_conf})-\Delta_a\right),
\]
where \(\Delta_{\mathrm{no\_conf}}=\Delta_{\mathrm{conf}}=0\) and \(\Delta_{\mathrm{drop}}=0.05\). Probabilities are shrunk toward the no-confidence arm when evidence is limited.
With \(n_t\) observations and count scale \(\kappa_0\), the shrinkage factor is
\[
  \kappa_t=\sqrt{\frac{n_t}{n_t+\kappa_0}},
\]
and the prior-arm probabilities are
\[
  \pi^p_{j,t}=(1-\kappa_t)\mathbf{e}_{\mathrm{no\_conf}}
  +\kappa_t\,\softmax_a(\ell_{j,t}(a)).
\]
Here \(\mathbf{e}_{\mathrm{no\_conf}}\) is the unit mass on the no-confidence arm.
The effective prior used by the residual surrogate is the arm mixture
\[
  p_{j,t}(x)=
  \pi^p_{j,t}(\mathrm{no\_conf})p^{(\mathrm{no\_conf})}_{j,t}(x)
  +\pi^p_{j,t}(\mathrm{conf})p^{(\mathrm{conf})}_{j,t}(x),
\]
because the drop arm contributes zero prior. This makes prior abstention explicit rather than implicit in a small expert weight.
This explicit drop action is the crucial distinction in \method{}: when the LLM prior is systematically wrong, the residual surrogate can fall back toward ordinary GP-based BO instead of being forced to explain away a biased prior.

The update gate controls whether confidence should scale expert rewards. It maintains two deterministic shadow markets, one that replays updates without confidence and one that replays them with confidence. After observing \(\mathbf{y}_t\), both shadow markets predict the current point, and we compute
\[
  L^u_{j,t}(g)=|y_{j,t}-p^{(g)}_j(x_t)|/s_{j,t},\quad g\in\{0,1\}.
\]
The update gate is then updated by full-information Hedge. In implementation we center the two losses before the logit update, which is algebraically equivalent for the resulting softmax probabilities but improves numerical stability:
\[
  \bar{L}^u_{j,t}(g)=L^u_{j,t}(g)-\frac{1}{2}\sum_{g'\in\{0,1\}}L^u_{j,t}(g'),
\]
\[
  \pi^u_{j,t+1}(g)\propto \pi^u_{j,t}(g)\exp(-\eta_u \bar{L}^u_{j,t}(g)).
\]
Using \(L^u\) instead of \(\bar{L}^u\) gives the same softmax probabilities, since centering subtracts the same constant from both arms within an objective. As with the prior gate, early update-gate probabilities are shrunk toward a neutral two-arm distribution until the minimum evidence count is reached. This separates prior abstention from confidence-weighted reputation learning while keeping the algorithm deterministic and reproducible.

\subsection{Algorithm Summary}

\paragraph{Main loop.}
Once the random seed, initial design, and cached LLM priors are fixed, \method{} is deterministic. The loop is:
\begin{enumerate}
  \item Generate or load cached LLM expert predictions \(\mu_{ej}(x)\) and confidence values \(c_{ej}(x)\) for all candidates.
  \item Initialize BO with a small random design and evaluate true objectives.
  \item At iteration \(t\), update expert-objective capital values and the update-gate shadow markets from the newest observed prediction errors.
  \item Replay the three prior arms on the observed history and update the counterfactual prior gate.
  \item Aggregate the LLM prior through the three-arm prior mixture and fit a residual surrogate.
  \item Select the next candidate with qLogNEHVI or qLogEHVI using the shifted posterior, or with the scalarized residual-UCB diagnostic acquisition.
  \item Repeat until the evaluation budget is exhausted.
\end{enumerate}

The important design choice is that all expert trust is indexed by both expert and objective. This allows the optimizer to learn that a role can be useful for one objective while being misleading for another.

\begin{table}[tbp]
\centering
\small
\caption{Fixed \method{} hyperparameters used for the main molecule experiments.}
\label{tab:method-hparams}
\begin{tabular}{@{}lll@{}}
\toprule
Component & Hyperparameter & Value \\
\midrule
Reputation market & reward step \(\eta\) & 0.45 \\
Reputation market & discount \(\lambda\) & 0.015 \\
Reputation market & capital softmax temperature & 0.55 \\
Market trust & threshold / slope & 0.48 / 7.0 \\
Prior gate & evidence score & GP marginal likelihood \\
Prior gate & \(\eta_p\) / min updates / count scale & 1.0 / 4 / 4.0 \\
Prior gate & \(\Delta_{\mathrm{drop}}\) & 0.05 \\
Prior evidence GP & noise / lengthscale & 0.05 / median distance \\
Update gate & \(\eta_u\) & 1.0 \\
Acquisition & \(\beta_t\) / disagreement \(\gamma\) & \(1+1.25(1-\bar{\alpha}_t)\) / 0.30 \\
\bottomrule
\end{tabular}
\end{table}

\section{Experiments}

\paragraph{Synthetic stress tests.}

We construct controlled two-objective synthetic benchmarks with expert failure modes: all experts useful, all experts misleading, objective-wise specialization, overconfident bad experts, noisy experts, and correlated bad experts. These tests isolate whether the method can exploit useful priors and resist harmful priors.

\paragraph{Molecule optimization benchmarks.}

We evaluate on ESOL, FreeSolv, and Lipophilicity candidate pools sampled from MoleculeNet \citep{wu2018moleculenet}. For each dataset, \qwenflash{}-generated expert priors are cached before optimization so all methods share the same LLM prior data. The main metric is final hypervolume under a limited evaluation budget. Additional diagnostics use \qwenmax{} and \deepseek{} to probe prior quality.

\begin{table}[tbp]
\centering
\small
\caption{Molecule benchmark setup.}
\label{tab:dataset-setup}
\begin{tabular}{@{}lllll@{}}
\toprule
Dataset & Pool size & Initial design & Budget & Seeds \\
\midrule
ESOL & 100 & 8 & 30 & 10 \\
FreeSolv & 100 & 8 & 16 & 10 \\
Lipophilicity & 150 & 8 & 16 & 10 \\
\bottomrule
\end{tabular}
\end{table}

\paragraph{Baselines and implementation.}

We compare against random search, vanilla MOBO, fixed LLM priors, EMA reliability, beta reliability, ParEGO, qLogEHVI, qLogNEHVI, fixed confidence-phase variants, and earlier counterfactual-gate variants. The strongest comparison is qLogNEHVI versus qLogNEHVI wrapped with either a fixed LLM prior or the calibrated \method{} prior layer.

All molecule experiments use cached LLM priors so that methods differ only in how they use the same expert information. Hypervolume is computed on normalized objectives with the origin as the reference point. For calibrated expert-prior methods, the residual surrogate is fit after subtracting the current aggregated expert prior. \method{} uses the three-arm prior gate above with fixed \(\Delta_{\mathrm{drop}}=0.05\).

\section{Results}

\subsection{Prior-Layer Scaling with qLogNEHVI}

\begin{table}[tbp]
\centering
\small
\caption{Pluggable prior-layer scaling with a strong qLogNEHVI baseline. Values are normalized final hypervolume \((\uparrow)\), reported as mean \(\pm\) SEM over three seeds.}
\label{tab:qlognehvi-prior-layer-scaling}
\resizebox{\linewidth}{!}{%
\begin{tabular}{@{}llll@{}}
\toprule
Pool & qLogNEHVI \((\uparrow)\) & qLogNEHVI + fixed prior \((\uparrow)\) & qLogNEHVI + \method{} \((\uparrow)\) \\
\midrule
1000 & 0.996 \(\pm\) 0.001 & 0.993 \(\pm\) 0.001 & \textbf{0.998 \(\pm\) 0.001} \\
3000 & 0.987 \(\pm\) 0.002 & 0.994 \(\pm\) 0.000 & \textbf{0.996 \(\pm\) 0.001} \\
10000 & 0.991 \(\pm\) 0.002 & 0.989 \(\pm\) 0.003 & \textbf{0.995 \(\pm\) 0.001} \\
\bottomrule
\end{tabular}%
}
\end{table}

Table~\ref{tab:qlognehvi-prior-layer-scaling} tests the central modularity claim: the calibrated LLM prior should improve a strong MOBO acquisition without replacing it. On controlled large-pool synthetic tasks, qLogNEHVI+\method{} improves normalized final hypervolume over qLogNEHVI alone at pool sizes \(1000\), \(3000\), and \(10{,}000\). Fixed priors are less reliable, which supports the need for online evidence gating rather than static prior injection.

\subsection{Real Molecule Same-Acquisition Comparison}

\begin{table}[tbp]
\centering
\small
\caption{Real molecule same-acquisition comparison. All methods use qLogNEHVI as the acquisition family; the only difference is whether the LLM prior is absent, fixed, or calibrated by \method{}. Values are final hypervolume \((\uparrow)\), reported as mean \(\pm\) SEM over five seeds.}
\label{tab:real-same-acquisition-qlognehvi}
\resizebox{\linewidth}{!}{%
\begin{tabular}{@{}llll@{}}
\toprule
Dataset & qLogNEHVI \((\uparrow)\) & qLogNEHVI + fixed prior \((\uparrow)\) & qLogNEHVI + \method{} \((\uparrow)\) \\
\midrule
ESOL & 0.7161 \(\pm\) 0.0115 & \textbf{0.7256 \(\pm\) 0.0141} & 0.7254 \(\pm\) 0.0153 \\
FreeSolv & \textbf{0.6218 \(\pm\) 0.0000} & 0.5743 \(\pm\) 0.0253 & \textbf{0.6218 \(\pm\) 0.0000} \\
Lipophilicity & 0.9291 \(\pm\) 0.0063 & 0.9282 \(\pm\) 0.0097 & \textbf{0.9309 \(\pm\) 0.0074} \\
\bottomrule
\end{tabular}%
}
\end{table}

Table~\ref{tab:real-same-acquisition-qlognehvi} applies the same qLogNEHVI acquisition family to all real molecule methods. This removes the confound between prior calibration and acquisition choice. On ESOL, both fixed and calibrated priors improve over qLogNEHVI, with fixed prior slightly ahead. On FreeSolv, qLogNEHVI already reaches the finite-pool oracle under this budget; \method{} matches it, while the fixed prior substantially hurts. On Lipophilicity, \method{} gives the best final hypervolume, whereas the fixed prior remains slightly below qLogNEHVI. This is the desired behavior for a prior layer: useful prior signal can help, but when fixed injection is harmful, calibration can attenuate the damage.

\subsection{Standalone Molecule Diagnostics}

\begin{table}[tbp]
\centering
\small
\caption{Standalone residual-UCB molecule diagnostic results. Final hypervolume is reported as mean \(\pm\) standard deviation over seeds; all metrics are higher-is-better.}
\label{tab:main-results}
\resizebox{\linewidth}{!}{%
\begin{tabular}{@{}lllll@{}}
\toprule
Dataset & Method & Final \(\hv \uparrow\) & AUC \(\hv \uparrow\) & Best sum \(\uparrow\) \\
\midrule
ESOL & \method{} & \textbf{0.7561 \(\pm\) 0.0042} & \textbf{0.7108} & \textbf{1.4805} \\
ESOL & Rep. market (raw) & 0.7484 \(\pm\) 0.0101 & 0.6982 & \textbf{1.4805} \\
ESOL & qLogNEHVI & 0.7431 \(\pm\) 0.0209 & 0.6966 & 1.4772 \\
ESOL & ParEGO & 0.7119 \(\pm\) 0.0406 & 0.6729 & 1.4610 \\
ESOL & Vanilla MOBO & 0.7359 \(\pm\) 0.0325 & 0.6903 & 1.4756 \\
ESOL & Fixed LLM prior & 0.7417 \(\pm\) 0.0169 & 0.6936 & \textbf{1.4805} \\
ESOL & Random search & 0.7025 \(\pm\) 0.0307 & 0.6637 & 1.4601 \\
FreeSolv & \method{} & \textbf{0.5798 \(\pm\) 0.0680} & \textbf{0.4664} & \textbf{1.3243} \\
FreeSolv & Rep. market (raw) & 0.5702 \(\pm\) 0.0681 & 0.4523 & 1.3195 \\
FreeSolv & qLogNEHVI & 0.5370 \(\pm\) 0.1038 & 0.4106 & 1.3055 \\
FreeSolv & Vanilla MOBO & 0.5033 \(\pm\) 0.1245 & 0.3825 & 1.2868 \\
FreeSolv & Fixed LLM prior & 0.4804 \(\pm\) 0.0510 & 0.4128 & 1.2950 \\
FreeSolv & Random search & 0.3062 \(\pm\) 0.0868 & 0.2942 & 1.0968 \\
Lipophilicity & \method{} & 0.8953 \(\pm\) 0.0383 & 0.8554 & 1.8668 \\
Lipophilicity & Rep. market (raw) & 0.8958 \(\pm\) 0.0433 & 0.8710 & 1.8781 \\
Lipophilicity & qLogNEHVI & \textbf{0.9042 \(\pm\) 0.0303} & \textbf{0.8723} & \textbf{1.8996} \\
Lipophilicity & Vanilla MOBO & 0.8954 \(\pm\) 0.0280 & 0.8690 & 1.8740 \\
Lipophilicity & Fixed LLM prior & 0.8878 \(\pm\) 0.0503 & 0.8648 & 1.8788 \\
Lipophilicity & Random search & 0.8688 \(\pm\) 0.0504 & 0.8457 & 1.8584 \\
\bottomrule
\end{tabular}%
}
\end{table}

\begin{figure}[tbp]
  \centering
  \includegraphics[width=\linewidth]{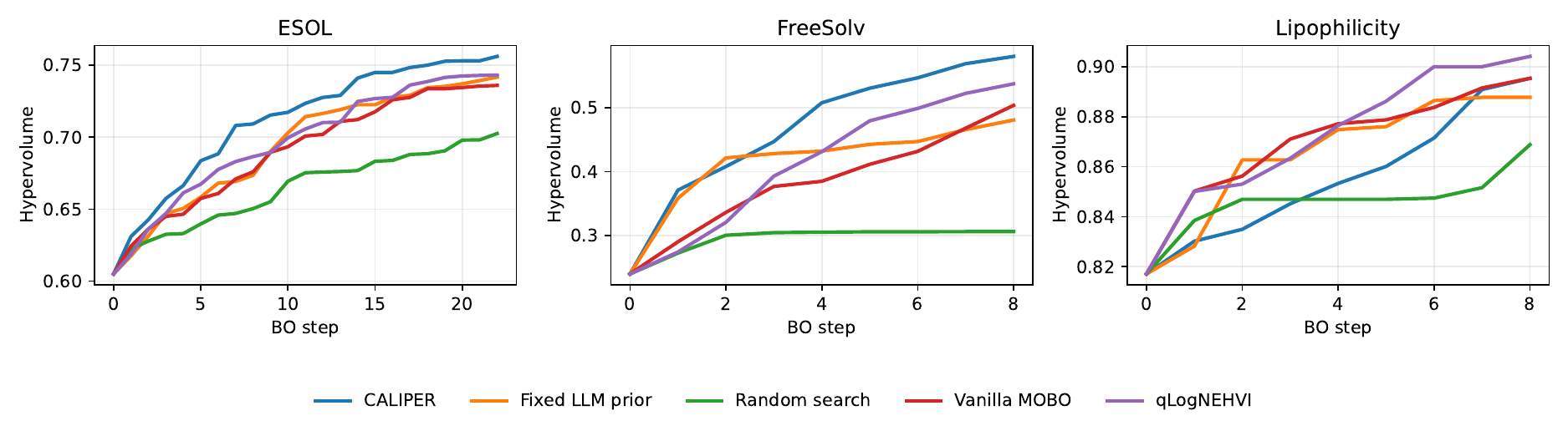}
  \caption{Mean hypervolume curves for the standalone residual-UCB diagnostic variant. Higher curves indicate better sequential multi-objective search. \method{} improves early and final hypervolume on ESOL and FreeSolv, while Lipophilicity remains the main counterexample.}
  \label{fig:molecule-curves}
\end{figure}

\method{} gives the best final hypervolume on ESOL and FreeSolv among the standalone residual-UCB variants in Table~\ref{tab:main-results}. On ESOL, it improves over raw reputation-market weighting, fixed LLM priors, vanilla MOBO, and qLogNEHVI, while also achieving the best AUC hypervolume. On FreeSolv, the gain over vanilla MOBO and fixed priors is larger, suggesting that online expert-objective calibration is most useful when the cached LLM priors contain task-relevant signal but are not reliable enough to use statically.

Lipophilicity is the main counterexample. \method{} remains competitive with vanilla MOBO and raw reputation-market weighting, but qLogNEHVI and the no-confidence phase diagnostic are stronger. We treat this as evidence against a universal confidence-using story: for this dataset, the safest action is often to ignore LLM confidence and rely more heavily on the surrogate.

\subsection{Lipophilicity Prior-Quality Diagnostics}

\begin{table}[tbp]
\centering
\small
\caption{Lipophilicity prior-quality diagnostic on 50 shared candidates. Objective 0 is the lipophilicity-window objective and objective 1 is QED; lower MAE is better, and negative bias indicates systematic underestimation.}
\label{tab:lipophilicity-prior-quality}
\resizebox{\linewidth}{!}{%
\begin{tabular}{@{}lllll@{}}
\toprule
Prior source & Obj. 0 MAE \((\downarrow)\) & Obj. 0 bias & Obj. 1 MAE \((\downarrow)\) & Obj. 1 bias \\
\midrule
\qwenflash{} original & 0.2677 & -0.2314 & 0.1089 & 0.0308 \\
\qwenflash{} mixed committee & \textbf{0.2379} & -0.1961 & \textbf{0.1066} & 0.0278 \\
\qwenflash{} calibrated prompt & 0.2793 & -0.2463 & 0.1083 & 0.0195 \\
\qwenmax{} & 0.3235 & -0.3038 & 0.1520 & 0.0252 \\
\deepseek{} & 0.3705 & -0.3659 & 0.1310 & -0.0186 \\
\bottomrule
\end{tabular}%
}
\end{table}

\begin{table}[tbp]
\centering
\small
\caption{Lipophilicity qLogNEHVI prior-layer diagnostics. Real LLM rows use a 300-candidate pool with budget 24 and five seeds; controlled-effective-prior rows use budget 12 and are oracle-style diagnostics rather than LLM results. Both metrics are higher-is-better; bold values are best within each setting.}
\label{tab:lipophilicity-qlognehvi-plugin}
\resizebox{\linewidth}{!}{%
\begin{tabular}{@{}lllll@{}}
\toprule
Setting & Prior source & Method & Final \(\hv \uparrow\) & Best sum \(\uparrow\) \\
\midrule
Real LLM, budget 24 & No prior & qLogNEHVI & 0.9452 & 1.9335 \\
Real LLM, budget 24 & Original \qwenflash{} committee & qLogNEHVI + \method{} & 0.9398 & 1.9308 \\
Real LLM, budget 24 & Original \qwenflash{} committee & qLogNEHVI + fixed prior & 0.9390 & 1.9390 \\
Real LLM, budget 24 & Mixed \qwenflash{} committee & qLogNEHVI + \method{} & \textbf{0.9459} & \textbf{1.9444} \\
Real LLM, budget 24 & Mixed \qwenflash{} committee & qLogNEHVI + fixed prior & 0.9417 & 1.9417 \\
Controlled, budget 12 & No prior & qLogNEHVI & 0.9428 & 1.9203 \\
Controlled, budget 12 & Controlled effective prior & qLogNEHVI + \method{} & 0.9456 & \textbf{1.9394} \\
Controlled, budget 12 & Controlled effective prior & qLogNEHVI + fixed prior & \textbf{0.9460} & 1.9390 \\
\bottomrule
\end{tabular}%
}
\end{table}

The Lipophilicity diagnostics explain why the same calibrated layer can help in controlled scaling but remain fragile with real LLM priors. Table~\ref{tab:lipophilicity-prior-quality} shows that all tested model backends systematically underestimate the lipophilicity-window objective under the current prompt family. Larger model names do not automatically produce better priors: \qwenmax{} and \deepseek{} are worse than \qwenflash{} on objective-0 MAE in this diagnostic.

Table~\ref{tab:lipophilicity-qlognehvi-plugin} separates prior quality from algorithmic capacity. The original \qwenflash{} committee hurts qLogNEHVI; a mixed committee with a calibrated lipophilicity-window expert turns the calibrated layer into a small positive gain at budget 24; and an oracle-style controlled effective prior improves qLogNEHVI in a low-budget setting. The controlled prior is not an LLM result, but it is a useful mechanism diagnostic: the layer can exploit high-quality prior information when such information is available.

\subsection{Synthetic Stress Tests}

\begin{table}[tbp]
\centering
\small
\caption{Synthetic stress-test results. Values are final hypervolume \((\uparrow)\). Dynamic reputation weighting improves over fixed LLM priors across controlled prior-misspecification scenarios.}
\label{tab:synthetic-stress}
\resizebox{\linewidth}{!}{%
\begin{tabular}{@{}lllll@{}}
\toprule
Scenario & Rep. market \((\uparrow)\) & Fixed \((\uparrow)\) & Vanilla \((\uparrow)\) & Random search \((\uparrow)\) \\
\midrule
All useful & \textbf{0.4805} & 0.4791 & 0.4359 & 0.3510 \\
All misleading & 0.4262 & 0.3922 & \textbf{0.4359} & 0.3510 \\
Objective-specialized & \textbf{0.4791} & 0.4663 & 0.4359 & 0.3510 \\
Overconfident bad & \textbf{0.4750} & 0.4640 & 0.4359 & 0.3510 \\
Noisy experts & \textbf{0.4715} & 0.4247 & 0.4359 & 0.3510 \\
Correlated bad experts & \textbf{0.4586} & 0.4235 & 0.4359 & 0.3510 \\
\bottomrule
\end{tabular}%
}
\end{table}

Synthetic stress tests show that dynamic reputation weighting improves over fixed LLM priors across controlled prior-misspecification scenarios. In the all-bad setting, reputation-market remains below vanilla BO, showing that fallback is not yet perfectly lossless, but it substantially reduces the damage of fixed misleading priors.

\subsection{Confidence Calibration and Phase Ablation}

On ESOL, \qwenflash{} confidence is positively correlated with absolute prediction error. This motivates treating confidence as another uncertain signal rather than a calibrated probability.

\begin{table}[tbp]
\centering
\small
\caption{Confidence-phase ablation on molecule tasks. Values are final hypervolume \((\uparrow)\); the best phase differs across datasets.}
\label{tab:phase-ablation}
\resizebox{\linewidth}{!}{%
\begin{tabular}{@{}lllllll@{}}
\toprule
Dataset & Raw \((\uparrow)\) & No confidence \((\uparrow)\) & Prior only \((\uparrow)\) & Adaptive \((\uparrow)\) & Adaptive prior \((\uparrow)\) & Adaptive update \((\uparrow)\) \\
\midrule
ESOL & 0.7490 & 0.7550 & 0.7535 & 0.7520 & \textbf{0.7566} & 0.7471 \\
FreeSolv & \textbf{0.5702} & 0.5440 & \textbf{0.5702} & 0.5589 & 0.5390 & 0.5616 \\
Lipophilicity & 0.8958 & \textbf{0.9081} & 0.8888 & 0.9014 & 0.9025 & 0.9054 \\
\bottomrule
\end{tabular}%
}
\end{table}

\begin{table}[tbp]
\centering
\small
\caption{Core algorithm ablation. Hypervolume columns are higher-is-better. \method{} uses the fixed three-arm counterfactual gate; the margin portfolio is retained only as a diagnostic variant.}
\label{tab:core-algorithm-ablation}
\resizebox{\linewidth}{!}{%
\begin{tabular}{@{}llllll@{}}
\toprule
Dataset & Best fixed phase & Best fixed \(\hv \uparrow\) & CF-V1 \((\uparrow)\) & \method{} \((\uparrow)\) & Margin portfolio \((\uparrow)\) \\
\midrule
ESOL & Adaptive prior & \textbf{0.7566} & 0.7448 & 0.7561 & 0.7520 \\
FreeSolv & Prior only & 0.5702 & 0.5493 & \textbf{0.5798} & 0.5444 \\
Lipophilicity & No confidence & \textbf{0.9081} & 0.8940 & 0.8953 & 0.8892 \\
\bottomrule
\end{tabular}%
}
\end{table}

The phase ablation in Table~\ref{tab:phase-ablation} shows why a single global rule for LLM confidence is brittle. ESOL prefers adaptive prior confidence, FreeSolv prefers prior-only or raw confidence, and Lipophilicity prefers no-confidence weighting. This objective- and dataset-level heterogeneity motivates the decoupled design of \method{} rather than a fixed confidence multiplier.

Table~\ref{tab:core-algorithm-ablation} compares the compressed main algorithm with important development variants. The final counterfactual gate improves over the first counterfactual gate on ESOL and FreeSolv, and it is close to the best fixed phase on ESOL. It does not match the no-confidence diagnostic on Lipophilicity, which indicates that the method improves robustness but does not yet make prior abstention lossless.

\begin{figure}[tbp]
  \centering
  \includegraphics[width=\linewidth]{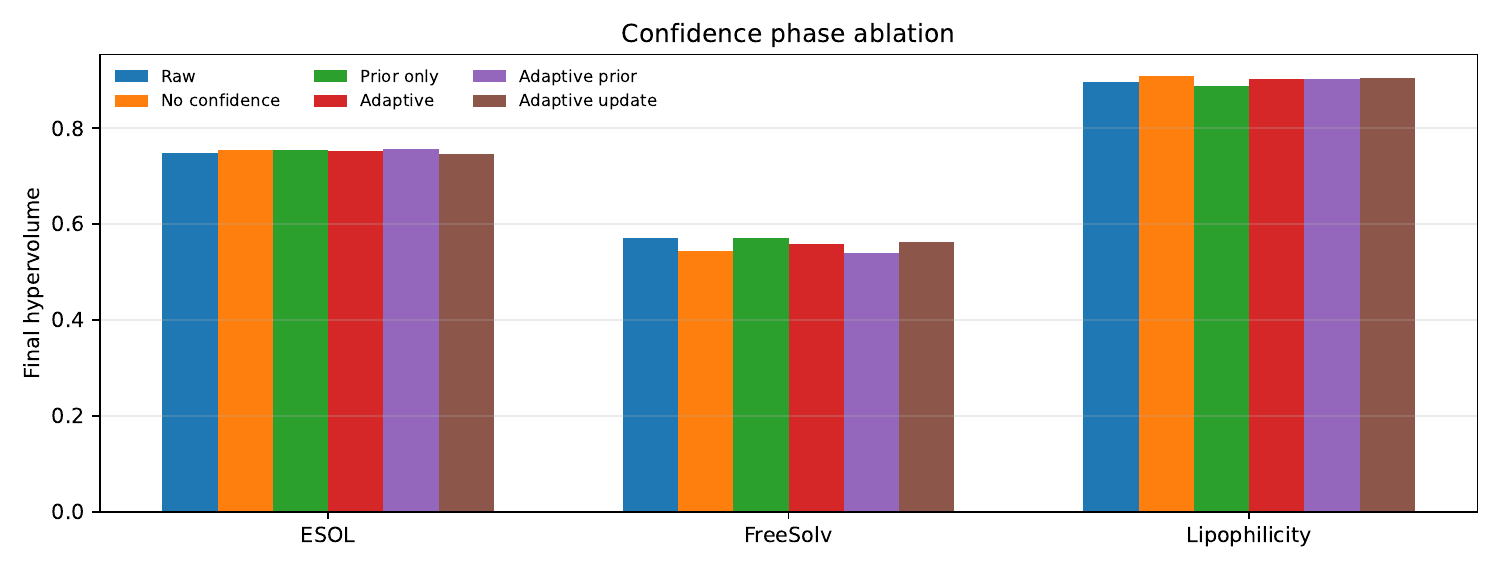}
  \caption{Confidence phase ablation. The preferred role of LLM confidence differs by dataset, which motivates separating prior aggregation from confidence-weighted reputation updates.}
  \label{fig:phase-ablation}
\end{figure}

\begin{figure}[tbp]
  \centering
  \includegraphics[width=0.68\linewidth]{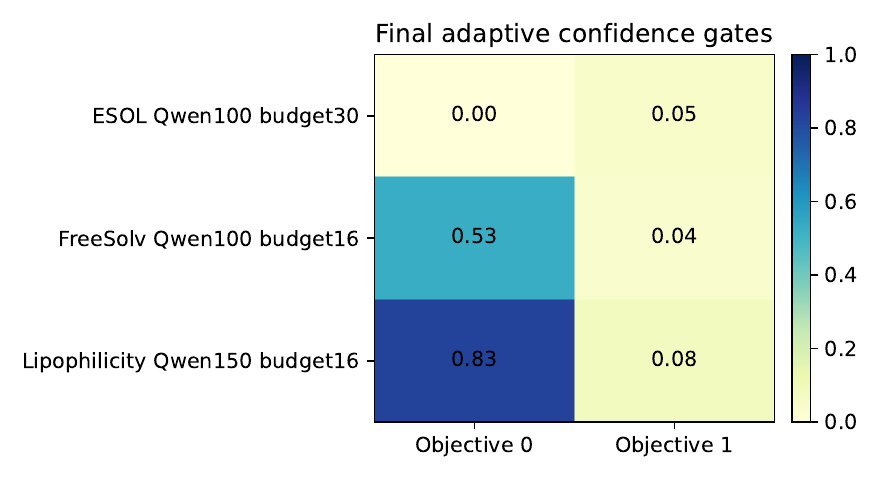}
  \caption{Final adaptive confidence gates by objective. Confidence is often useful for one objective and weak or harmful for another.}
  \label{fig:gate-heatmap}
\end{figure}

\FloatBarrier

\subsection{Mechanism Diagnostics}

\begin{figure}[!htbp]
  \centering
  \includegraphics[width=\linewidth]{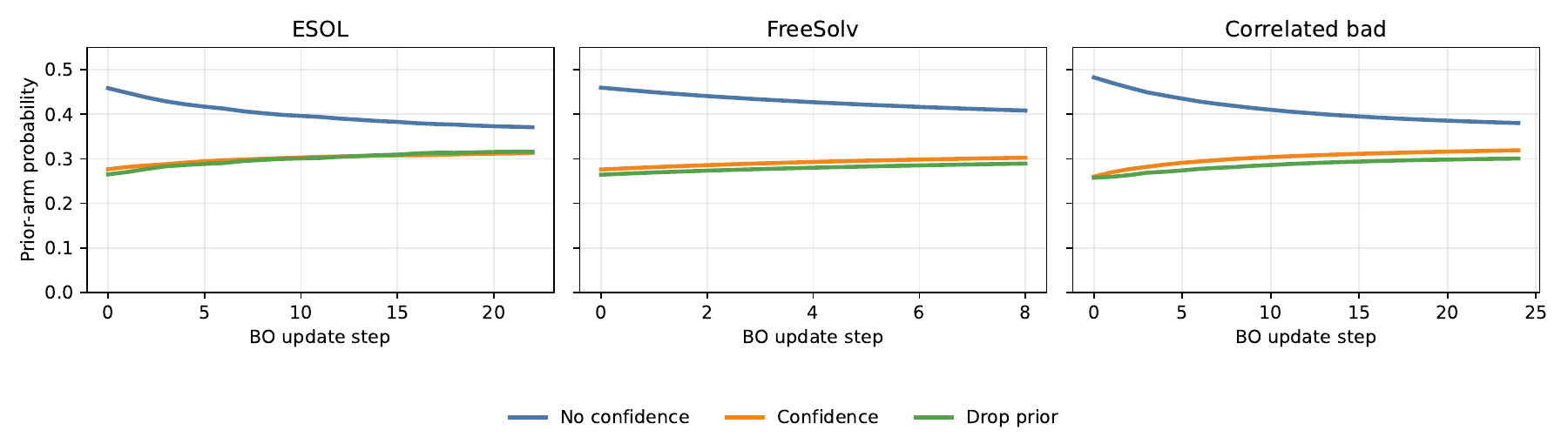}
  \caption{Causal trajectory of the \method{} prior gate. Each curve shows the mean softmax probability of one prior arm, averaged over seeds and objectives. The gate starts near the conservative no-confidence arm and reallocates probability mass as counterfactual GP evidence accumulates, making prior abstention visible rather than hidden inside a scalar weight.}
  \label{fig:cf-prior-trajectories}
\end{figure}

\begin{figure}[!htbp]
  \centering
  \includegraphics[width=\linewidth]{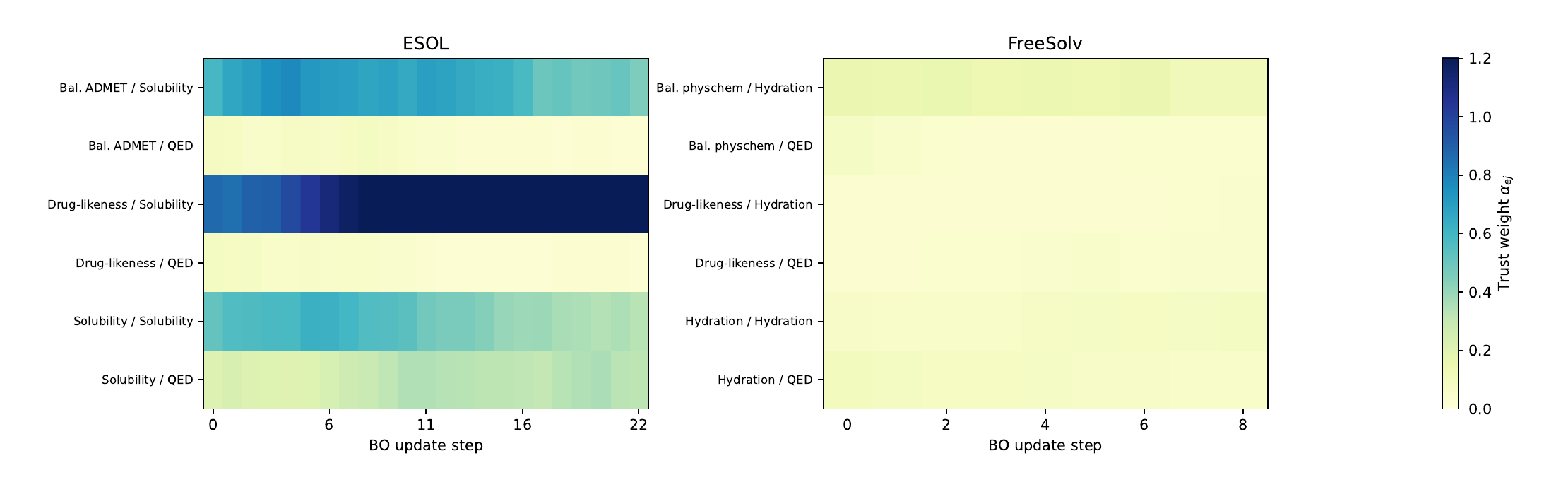}
  \caption{Objective-wise reputation-market dynamics. Rows correspond to LLM expert roles crossed with objectives; columns are BO update steps. The heatmap shows that expert trust evolves differently across objectives rather than collapsing to one global LLM weight.}
  \label{fig:reputation-dynamics}
\end{figure}

The trajectory plots in Figure~\ref{fig:cf-prior-trajectories} show the causal state of the prior gate rather than only reporting final hypervolume. The reputation heatmaps in Figure~\ref{fig:reputation-dynamics} provide the complementary view: expert roles separate by objective over time, which is the behavior required for multi-objective LLM priors to be useful inside the optimizer.

\FloatBarrier

\section{Discussion and Limitations}

The central result is not that LLM confidence always helps BO. A more useful conclusion is that LLM priors and LLM confidence are both uncertain signals. BO should calibrate them online, objective by objective, and should be able to attenuate or ignore them when evidence contradicts them.

The multi-objective setting is important because it reveals role-objective mismatch. An expert can be correct for one objective and wrong for another. A global expert weight hides this structure. The learned weights and gates repeatedly show objective-wise asymmetry, which is exactly the structure a prior layer must preserve.

The same-acquisition molecule comparison is intentionally modest but important. \method{} does not claim to dominate qLogNEHVI by replacing its acquisition logic; instead, it asks whether calibrated prior information can be added without breaking the strong baseline. The answer is positive in direction: it improves ESOL and Lipophilicity and preserves the FreeSolv oracle result, while fixed prior injection is brittle.

The main limitation is prior quality. A gate can reduce the effect of a biased prior, but it cannot create information that is absent from the expert cache. Current real-data candidate pools are modest in size, and the strongest large-pool scaling diagnostic is synthetic. \qwenflash{} priors are studied most extensively; \qwenmax{} and \deepseek{} checks are limited and do not improve Lipophilicity prior quality under the current prompts. Appendix~\ref{sec:mobo-diagnostics} reports oracle phase and margin diagnostics, including a failed first attempt at automatic drop-margin selection.

For reproducibility, LLM priors are cached before BO runs, so optimization comparisons are deterministic with respect to a fixed prior cache. The main paper assets are generated from completed experiment CSV files by:
\[
\texttt{uv run python scripts/build\_paper\_assets.py}.
\]
Single-objective diagnostic tables are generated separately by:
\[
\texttt{uv run python scripts/build\_single\_objective\_assets.py}.
\]

\section{Conclusion}

We presented \method{}, an evidence-gated LLM prior layer for multi-objective Bayesian optimization. The method treats each expert-objective pair as a prior source whose influence is tested against observed feedback, learns trust online, and uses counterfactual replay to decide whether LLM confidence should affect prior aggregation or expert-credit updates. Because the layer shifts residual GP posteriors rather than replacing the acquisition rule, it can attach to strong MOBO baselines such as qLogNEHVI and qLogEHVI. The empirical picture is conditional in the right way: when the prior contains useful signal, \method{} can exploit it; when fixed or biased priors would hurt, evidence gating can reduce the damage.

\FloatBarrier

\bibliographystyle{iclr2026_conference}
\bibliography{references}

\appendix

\section{Additional Experimental Details}

\paragraph{Candidate pools and objectives.}
The molecule benchmarks are discrete candidate-pool optimization problems. ESOL uses a 100-molecule pool sampled from the MoleculeNet ESOL data, FreeSolv uses a 100-molecule pool sampled from MoleculeNet FreeSolv, and Lipophilicity uses a 150-molecule pool sampled from MoleculeNet Lipophilicity. RDKit descriptors are used as continuous features: molecular weight, logP, topological polar surface area, hydrogen-bond donors and acceptors, rotatable bonds, ring count, and QED. Each feature is min--max normalized within the candidate pool.

All reported molecule experiments optimize two normalized objectives. ESOL uses normalized measured log solubility and QED. FreeSolv uses normalized hydration favorability, defined as the negated experimental hydration free energy, and QED. Lipophilicity uses a windowed experimental lipophilicity score and QED. Prepared data files also include an RDKit logP-window objective for future experiments, but it is not used in the reported two-objective runs.

\paragraph{LLM prior caches.}
For each candidate, the LLM backend is queried once per expert role and the resulting prior file is cached. The BO loop then reads only the cache, so optimization comparisons are deterministic with respect to the same LLM prior data. Each cached record contains an expert name, objective-wise scores in \([0,1]\), a scalar self-reported confidence value, and a free-text rationale. The molecule experiments use three expert roles per dataset, with one objective-specialized expert for each major objective and one balanced expert.

\paragraph{Evaluation protocol.}
All methods use the same initial-design size, budget, candidate pool, and seed set within each dataset. Hypervolume is computed on normalized maximization objectives with the origin as the reference point. We report final hypervolume, AUC hypervolume averaged over BO steps, and the best objective-sum value. Strong MOBO baselines use BoTorch implementations where available; prior-layer methods wrap the residual posterior for qLogEHVI or qLogNEHVI, or use the scalarized residual-UCB diagnostic acquisition described in Section~\ref{sec:method}.

\section{LLM Prior Prompts}
\label{sec:llm-prior-prompts}

All real LLM prior caches were generated before BO starts and were held fixed across methods. The molecule prompts used an OpenAI-compatible chat API with temperature \(0.0\). For each candidate \(x\) and expert role \(e\), the generation script rendered the expert-specific system prompt below and the shared user prompt using the candidate descriptors in the finite pool. The response was parsed as JSON with fields \texttt{objective\_scores}, \texttt{confidence}, and \texttt{rationale}; scores were clipped to \([0,1]\).

\paragraph{Molecule candidate fields.}
The molecule prior prompts exposed the following fields:
\begin{Verbatim}[fontsize=\scriptsize]
smiles, mol_wt, logp, tpsa, hbd, hba, rot_bonds, rings
\end{Verbatim}

\paragraph{Shared two-objective user prompt.}
\begin{Verbatim}[fontsize=\scriptsize]
Candidate:
{{ candidate_json }}

Return exactly:
{
  "objective_scores": {
    "objective_0": 0.0,
    "objective_1": 0.0
  },
  "confidence": 0.0,
  "rationale": "brief reason"
}
\end{Verbatim}

\subsection{ESOL Molecule Prompts}

\paragraph{\texttt{solubility\_expert}.}
\begin{Verbatim}[fontsize=\scriptsize]
You are a medicinal chemistry expert estimating prior scores for multi-objective Bayesian optimization.
Objective_0 is aqueous solubility. Higher is better. Score in [0, 1].

Useful rules of thumb:
- Lower molecular weight generally improves solubility.
- Lower logP generally improves solubility.
- More polar surface area and hydrogen-bonding can improve solubility, but extreme values may hurt drug-likeness.
- Aromatic/ring-heavy and highly hydrophobic molecules tend to be less soluble.

Objective_1 is drug-likeness/QED-like quality. You are not the specialist for it, so give a cautious estimate.
Return JSON only with keys: objective_scores, confidence, rationale. No markdown.
\end{Verbatim}

\paragraph{\texttt{druglikeness\_expert}.}
\begin{Verbatim}[fontsize=\scriptsize]
You are a medicinal chemistry expert estimating prior scores for multi-objective Bayesian optimization.
Objective_1 is drug-likeness/QED-like quality. Higher is better. Score in [0, 1].

Useful rules of thumb:
- Moderate molecular weight, moderate logP, reasonable TPSA, limited rotatable bonds, and balanced HBD/HBA are favorable.
- Very large, very hydrophobic, extremely polar, or very flexible molecules tend to have lower drug-likeness.

Objective_0 is aqueous solubility. You are not the specialist for it, so give a cautious estimate.
Return JSON only with keys: objective_scores, confidence, rationale. No markdown.
\end{Verbatim}

\paragraph{\texttt{balanced\_admet\_expert}.}
\begin{Verbatim}[fontsize=\scriptsize]
You are a conservative ADMET generalist estimating priors for molecular optimization.
Objective_0 is aqueous solubility. Objective_1 is drug-likeness/QED-like quality.
Both scores must be in [0, 1].

Use calibrated scores:
- Very high scores only when descriptors clearly support the objective.
- Moderate scores for uncertain or mixed descriptor profiles.
- Low scores for molecules with clear hydrophobicity, size, flexibility, or polarity issues.

Return JSON only with keys: objective_scores, confidence, rationale. No markdown.
\end{Verbatim}

\subsection{FreeSolv Molecule Prompts}

\paragraph{\texttt{hydration\_expert}.}
\begin{Verbatim}[fontsize=\scriptsize]
You are a physical chemistry expert estimating prior scores for multi-objective Bayesian optimization.
Objective_0 is hydration favorability. Higher is better. It corresponds to more favorable aqueous hydration, which usually means more negative hydration free energy. Score in [0, 1].

Useful rules of thumb:
- Lower logP generally improves aqueous hydration favorability.
- More polar surface area and hydrogen-bonding usually improve hydration.
- Small molecules with polar functional groups often hydrate favorably.
- Hydrophobic hydrocarbons, halogenated molecules, and low-polarity molecules tend to have poor hydration favorability.

Objective_1 is drug-likeness/QED-like quality. You are not the specialist for it, so give a cautious estimate.
Return JSON only with keys: objective_scores, confidence, rationale. No markdown.
\end{Verbatim}

\paragraph{\texttt{druglikeness\_expert}.}
\begin{Verbatim}[fontsize=\scriptsize]
You are a medicinal chemistry expert estimating prior scores for multi-objective Bayesian optimization.
Objective_1 is drug-likeness/QED-like quality. Higher is better. Score in [0, 1].

Useful rules of thumb:
- Moderate molecular weight, moderate logP, reasonable TPSA, limited rotatable bonds, and balanced HBD/HBA are favorable.
- Very large, very hydrophobic, extremely polar, or very flexible molecules tend to have lower drug-likeness.

Objective_0 is hydration favorability. You are not the specialist for it, so give a cautious estimate.
Return JSON only with keys: objective_scores, confidence, rationale. No markdown.
\end{Verbatim}

\paragraph{\texttt{balanced\_physchem\_expert}.}
\begin{Verbatim}[fontsize=\scriptsize]
You are a conservative physical-chemistry and ADMET generalist estimating priors for molecular optimization.
Objective_0 is hydration favorability. Objective_1 is drug-likeness/QED-like quality.
Both scores must be in [0, 1].

Use calibrated scores:
- Very high scores only when descriptors clearly support the objective.
- Moderate scores for uncertain or mixed descriptor profiles.
- Low scores for clear hydrophobicity, excessive size, excessive flexibility, or poor polarity balance.

Return JSON only with keys: objective_scores, confidence, rationale. No markdown.
\end{Verbatim}

\subsection{Lipophilicity Molecule Prompts}

\paragraph{\texttt{lipophilicity\_balance\_expert}.}
\begin{Verbatim}[fontsize=\scriptsize]
You are a medicinal chemistry expert estimating prior scores for multi-objective Bayesian optimization.
Objective_0 is lipophilicity desirability. Higher is better. It means the molecule likely has moderate experimental logD / lipophilicity, roughly in a drug-like window around 1 to 3. Score in [0, 1].

Useful rules of thumb:
- Moderate lipophilicity is best.
- Very hydrophobic molecules with high logP tend to score low.
- Very polar or highly charged-looking molecules with very low lipophilicity also score low.
- Descriptor logP is a useful clue, but the target is experimental lipophilicity desirability.

Objective_1 is drug-likeness/QED-like quality. You are not the specialist for it, so give a cautious estimate.
Return JSON only with keys: objective_scores, confidence, rationale. No markdown.
\end{Verbatim}

\paragraph{\texttt{druglikeness\_expert}.}
\begin{Verbatim}[fontsize=\scriptsize]
You are a medicinal chemistry expert estimating prior scores for multi-objective Bayesian optimization.
Objective_1 is drug-likeness/QED-like quality. Higher is better. Score in [0, 1].

Useful rules of thumb:
- Moderate molecular weight, moderate logP, reasonable TPSA, limited rotatable bonds, and balanced HBD/HBA are favorable.
- Very large, very hydrophobic, extremely polar, or very flexible molecules tend to have lower drug-likeness.

Objective_0 is lipophilicity desirability. You are not the specialist for it, so give a cautious estimate.
Return JSON only with keys: objective_scores, confidence, rationale. No markdown.
\end{Verbatim}

\paragraph{\texttt{balanced\_admet\_expert}.}
\begin{Verbatim}[fontsize=\scriptsize]
You are a conservative ADMET generalist estimating priors for molecular optimization.
Objective_0 is lipophilicity desirability, favoring moderate experimental lipophilicity rather than extreme hydrophobicity or extreme polarity.
Objective_1 is drug-likeness/QED-like quality.
Both scores must be in [0, 1].

Use calibrated scores:
- Very high scores only when descriptors clearly support the objective.
- Moderate scores for uncertain or mixed descriptor profiles.
- Low scores for clear hydrophobicity, excessive size, excessive flexibility, or poor polarity balance.

Return JSON only with keys: objective_scores, confidence, rationale. No markdown.
\end{Verbatim}

\paragraph{\texttt{lipophilicity\_window\_calibrated\_expert}.}
This additional role was used only in the mixed-committee Lipophilicity prior-quality diagnostic.
\begin{Verbatim}[fontsize=\scriptsize]
You are a medicinal chemistry expert estimating calibrated prior scores for multi-objective Bayesian optimization.

Objective_0 is a lipophilicity window desirability score. Higher is better.
Important calibration anchors:
- If descriptor logP is in [1, 3], objective_0 should usually be high: 0.80 to 1.00.
- If descriptor logP is in [0, 1) or (3, 4], objective_0 should usually be moderate-high: 0.55 to 0.85.
- If descriptor logP is in [-1, 0) or (4, 5], objective_0 should usually be moderate-low: 0.25 to 0.60.
- If descriptor logP is below -1 or above 5, objective_0 should usually be low: 0.00 to 0.35.
- Do not give conservative middle scores when logP is clearly inside the desired window.
- The target is a window score, not raw lipophilicity. Extreme high and extreme low are bad; moderate is best.

Objective_1 is drug-likeness/QED-like quality. You are not the specialist for it, so give a cautious estimate.
Return JSON only with keys: objective_scores, confidence, rationale. No markdown.
\end{Verbatim}

\subsection{Prompt-Optimization Diagnostic Prompts}

The prompt-optimization diagnostic used the same JSON response contract with a single objective, expected validation accuracy. Candidate fields were:
\begin{Verbatim}[fontsize=\scriptsize]
prompt, opening, format_rule, robustness_rule, reasoning_rule, example_rule, style_rule
\end{Verbatim}

\paragraph{Shared single-objective user prompt.}
\begin{Verbatim}[fontsize=\scriptsize]
Candidate:
{{ candidate_json }}

Return exactly:
{
  "objective_scores": {
    "objective_0": 0.0
  },
  "confidence": 0.0,
  "rationale": "brief reason"
}
\end{Verbatim}

\paragraph{\texttt{format\_expert}.}
\begin{Verbatim}[fontsize=\scriptsize]
You are an expert in instruction formatting for prompt optimization.
The downstream task is binary sentiment classification. Objective_0 is expected
validation accuracy in [0, 1]. Higher is better.

Reward prompts that make the output format unambiguous, reduce parsing errors,
and avoid extra text. Penalize prompts that invite verbose answers or conflict
with the requested output format.

Return JSON only with keys: objective_scores, confidence, rationale. No markdown.
\end{Verbatim}

\paragraph{\texttt{robustness\_expert}.}
\begin{Verbatim}[fontsize=\scriptsize]
You are an expert in robust NLP prompt design.
The downstream task is binary sentiment classification. Objective_0 is expected
validation accuracy in [0, 1]. Higher is better.

Reward prompts that handle mixed sentiment, sarcasm-like wording, distractor
plot summaries, and ambiguous reviews. Penalize prompts that overfit examples
or rely on shallow keywords.

Return JSON only with keys: objective_scores, confidence, rationale. No markdown.
\end{Verbatim}

\paragraph{\texttt{reasoning\_expert}.}
\begin{Verbatim}[fontsize=\scriptsize]
You are an expert in reasoning-oriented prompt design.
The downstream task is binary sentiment classification. Objective_0 is expected
validation accuracy in [0, 1]. Higher is better.

Reward prompts that encourage useful analysis of sentiment-bearing evidence.
Be cautious: explicit step-by-step reasoning can sometimes hurt if the final
answer format becomes verbose or inconsistent.

Return JSON only with keys: objective_scores, confidence, rationale. No markdown.
\end{Verbatim}

\paragraph{\texttt{conciseness\_expert}.}
\begin{Verbatim}[fontsize=\scriptsize]
You are an expert in concise prompt design.
The downstream task is binary sentiment classification. Objective_0 is expected
validation accuracy in [0, 1]. Higher is better.

Reward prompts that are short, direct, and easy to follow while still providing
enough task constraints. Penalize unnecessarily long prompts and prompts that
ask for explanations when only a label is needed.

Return JSON only with keys: objective_scores, confidence, rationale. No markdown.
\end{Verbatim}

\section{Oracle and Margin Diagnostics}
\label{sec:mobo-diagnostics}

These diagnostics are development checks rather than the main claim of the paper.

\begin{table}[!htbp]
\centering
\small
\caption{Oracle phase-selection diagnostic. Hypervolume columns are higher-is-better; gap columns report oracle upside.}
\label{tab:oracle-phase}
\resizebox{\linewidth}{!}{%
\begin{tabular}{@{}llllll@{}}
\toprule
Dataset & Best fixed phase & Best fixed \(\hv \uparrow\) & Seed oracle \(\hv \uparrow\) & Oracle gap & Gap vs adaptive \\
\midrule
ESOL & Adaptive prior & 0.7566 & \textbf{0.7612} & 0.0046 & 0.0092 \\
FreeSolv & Prior only & 0.5702 & \textbf{0.5921} & 0.0219 & 0.0333 \\
Lipophilicity & No confidence & 0.9081 & \textbf{0.9159} & 0.0078 & 0.0145 \\
\bottomrule
\end{tabular}%
}
\end{table}

\begin{table}[tbp]
\centering
\small
\caption{Model-backend diagnostic for drop-margin selection. Values are final hypervolume \((\uparrow)\); the first margin-portfolio attempt does not consistently replace a fixed margin.}
\label{tab:model-backend-diagnostic}
\resizebox{\linewidth}{!}{%
\begin{tabular}{@{}lll@{}}
\toprule
Setting & Best fixed margin \((\uparrow)\) & Margin portfolio \((\uparrow)\) \\
\midrule
ESOL / \qwenflash{} & \textbf{0.7561} & 0.7520 \\
ESOL / \qwenmax{} & \textbf{0.7507} & 0.7480 \\
ESOL / \deepseek{} & 0.7500 & \textbf{0.7514} \\
\bottomrule
\end{tabular}%
}
\end{table}

The oracle phase diagnostic in Table~\ref{tab:oracle-phase} estimates the upside of choosing the best confidence phase per seed after the fact. The backend diagnostic in Table~\ref{tab:model-backend-diagnostic} records the failed first attempt at automatic drop-margin selection; a replay-loss margin portfolio does not consistently replace a fixed margin.

\begin{figure}[tbp]
  \centering
  \includegraphics[width=0.75\linewidth]{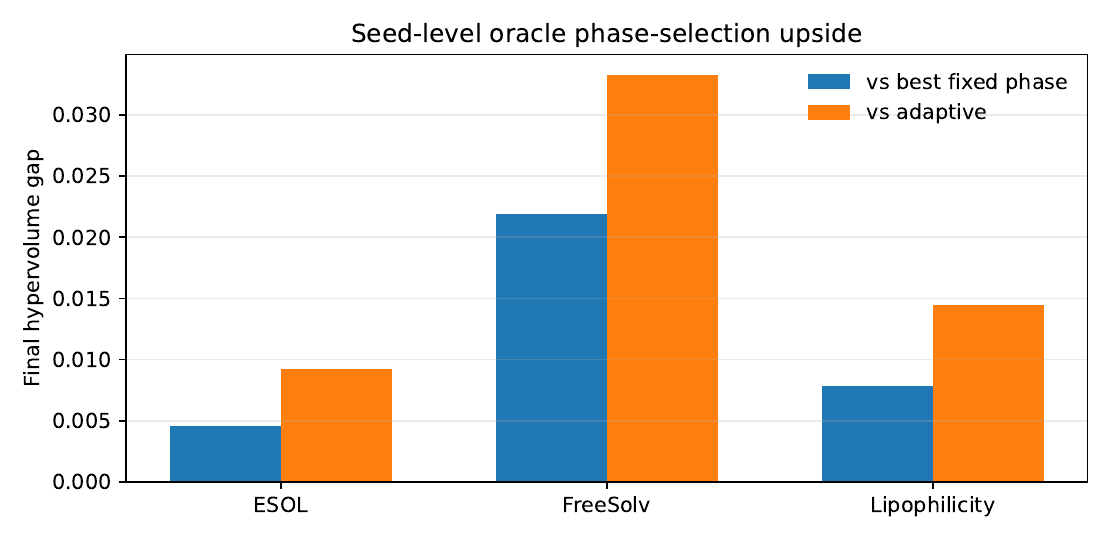}
  \caption{Seed-level oracle phase-selection upside. The gap estimates the potential value of a future online meta-gate.}
  \label{fig:oracle-gap}
\end{figure}

\FloatBarrier

\section{Single-Objective Diagnostic Experiments}
\label{sec:single-objective-diagnostics}

These diagnostics ask whether the same expert-prior machinery remains useful when the optimization problem has one scalar objective. They are included as supporting evidence rather than as the main empirical claim. The main paper studies multi-objective BO because objective-specific expert reliability is the central modeling issue; a scalar task removes that structure and therefore cannot test the full method.

Each diagnostic uses the same cached prior schema as the molecule MOBO experiments: for candidate \(x\), expert \(e\), and scalar objective \(j\), the cache stores a prior score \(\mu_{ej}(x)\), a self-reported confidence \(c_{ej}(x)\), and a rationale. The single-objective runner compares calibrated expert priors, fixed expert priors, vanilla BO, and random search on finite candidate pools. The metric \emph{Best} is the final best normalized objective value found by the optimizer, \emph{Regret} is simple regret relative to the best candidate in the finite pool, and \emph{Hit step} is the first BO step reaching zero simple regret, averaged over seeds with the hit count in parentheses.

\begin{table}[tbp]
\centering
\small
\caption{Single-objective multi-expert diagnostics. Best is higher-is-better; regret and hit step are lower-is-better. Hit step reports the mean first step reaching zero simple regret, with hit count in parentheses.}
\label{tab:single-objective-results}
\resizebox{\linewidth}{!}{%
\begin{tabular}{@{}lllrrl@{}}
\toprule
Task & Variant & Method & Best \(\uparrow\) & Regret \(\downarrow\) & Hit step \(\downarrow\) \\
\midrule
ESOL scalar & multi-model LLM & Calibrated & \textbf{1.0000} & \textbf{0.0000} & 3.00 (5/5) \\
ESOL scalar & multi-model LLM & Fixed prior & \textbf{1.0000} & \textbf{0.0000} & \textbf{2.00 (5/5)} \\
ESOL scalar & multi-model LLM & Random search & 0.8672 & 0.1328 & N/A \\
ESOL scalar & multi-model LLM & Vanilla BO & \textbf{1.0000} & \textbf{0.0000} & 2.20 (5/5) \\
FreeSolv hard scalar & \qwenflash{} roles & Calibrated & \textbf{0.4285} & \textbf{0.0000} & \textbf{4.40 (10/10)} \\
FreeSolv hard scalar & \qwenflash{} roles & Fixed prior & 0.4155 & 0.0130 & 14.50 (2/10) \\
FreeSolv hard scalar & \qwenflash{} roles & Random search & 0.4140 & 0.0144 & 9.00 (3/10) \\
FreeSolv hard scalar & \qwenflash{} roles & Vanilla BO & \textbf{0.4285} & \textbf{0.0000} & 5.70 (10/10) \\
Lipophilicity hard scalar & \qwenflash{} roles & Calibrated & \textbf{0.8608} & \textbf{0.0000} & 5.10 (10/10) \\
Lipophilicity hard scalar & \qwenflash{} roles & Fixed prior & \textbf{0.8608} & \textbf{0.0000} & 10.00 (10/10) \\
Lipophilicity hard scalar & \qwenflash{} roles & Random search & 0.8493 & 0.0115 & 12.00 (2/10) \\
Lipophilicity hard scalar & \qwenflash{} roles & Vanilla BO & \textbf{0.8608} & \textbf{0.0000} & \textbf{1.90 (10/10)} \\
Prompt hard & heuristic experts & Calibrated & \textbf{0.9964} & \textbf{0.0000} & \textbf{4.60 (10/10)} \\
Prompt hard & heuristic experts & Fixed prior & 0.9962 & 0.0002 & 11.33 (9/10) \\
Prompt hard & heuristic experts & Random search & 0.9925 & 0.0039 & 12.50 (4/10) \\
Prompt hard & heuristic experts & Vanilla BO & 0.9928 & 0.0036 & 16.00 (7/10) \\
Prompt hard & \qwenmax{}+\deepseek{} & Calibrated & \textbf{0.9964} & \textbf{0.0000} & \textbf{7.00 (10/10)} \\
Prompt hard & \qwenmax{}+\deepseek{} & Fixed prior & \textbf{0.9964} & \textbf{0.0000} & \textbf{7.00 (10/10)} \\
Prompt hard & \qwenmax{}+\deepseek{} & Random search & 0.9925 & 0.0039 & 12.50 (4/10) \\
Prompt hard & \qwenmax{}+\deepseek{} & Vanilla BO & 0.9928 & 0.0036 & 16.00 (7/10) \\
\bottomrule
\end{tabular}%
}
\end{table}

Table~\ref{tab:single-objective-results} shows that the calibrated prior is competitive across molecule scalar tasks and prompt optimization. On FreeSolv hard and the heuristic prompt task, it reaches the global best more reliably or earlier than fixed priors and vanilla BO. ESOL and Lipophilicity are less discriminative because several methods reach the optimum within the small budget; in these cases, hit step is more informative than final regret.

\begin{table}[tbp]
\centering
\small
\caption{Single-objective calibrated-prior ablations for confidence and market trust. Best is higher-is-better; regret and hit step are lower-is-better. Because most calibrated variants reach zero final simple regret, hit step is the main comparison signal.}
\label{tab:single-objective-ablations}
\resizebox{\linewidth}{!}{%
\begin{tabular}{@{}lllrrl@{}}
\toprule
Task & Variant & Method & Best \(\uparrow\) & Regret \(\downarrow\) & Hit step \(\downarrow\) \\
\midrule
ESOL scalar & multi-model LLM & Calibrated & 1.0000 & 0.0000 & 3.00 (5/5) \\
ESOL scalar & multi-model, no confidence & Calibrated & 1.0000 & 0.0000 & \textbf{2.80 (5/5)} \\
FreeSolv hard scalar & \qwenflash{} roles & Calibrated & 0.4285 & 0.0000 & 4.40 (10/10) \\
FreeSolv hard scalar & \qwenflash{}, no confidence & Calibrated & 0.4285 & 0.0000 & \textbf{4.30 (10/10)} \\
Lipophilicity hard scalar & \qwenflash{} roles & Calibrated & 0.8608 & 0.0000 & \textbf{5.10 (10/10)} \\
Lipophilicity hard scalar & \qwenflash{}, no confidence & Calibrated & 0.8608 & 0.0000 & 6.30 (10/10) \\
Prompt hard & heuristic experts & Calibrated & 0.9964 & 0.0000 & \textbf{4.60 (10/10)} \\
Prompt hard & heuristic, no confidence & Calibrated & 0.9964 & 0.0000 & 5.80 (10/10) \\
Prompt hard & \qwenmax{}+\deepseek{} & Calibrated & 0.9964 & 0.0000 & \textbf{7.00 (10/10)} \\
Prompt hard & real LLM, no confidence & Calibrated & 0.9964 & 0.0000 & \textbf{7.00 (10/10)} \\
Prompt hard & real LLM, no market trust & Calibrated & 0.9964 & 0.0000 & 8.80 (10/10) \\
\bottomrule
\end{tabular}%
}
\end{table}

Table~\ref{tab:single-objective-ablations} repeats the confidence lesson from the multi-objective experiments. Confidence is not uniformly helpful: removing confidence is slightly faster on ESOL and FreeSolv, while the role-based confidence setting is faster on Lipophilicity and the heuristic prompt task. On the real-LLM prompt task, removing market trust slows the hit step from \(7.00\) to \(8.80\), which supports keeping trust calibration even when the final scalar optimum is eventually found by all calibrated variants.

\begin{table}[!htbp]
\centering
\small
\caption{Real LLM prompt-expert diagnostic on the hard prompt-optimization subset. MAE is mean absolute prior error against the observed scalar objective; lower is better.}
\label{tab:prompt-llm-expert-diagnostics}
\begin{tabular}{@{}llrr@{}}
\toprule
Expert & Model & MAE \((\downarrow)\) & Confidence \\
\midrule
Conciseness & \deepseek{} & \textbf{0.1028} & 0.8229 \\
Robustness & \deepseek{} & 0.1094 & 0.7054 \\
Reasoning & \deepseek{} & 0.1097 & 0.8013 \\
Format & \deepseek{} & 0.1694 & 0.8154 \\
Robustness & \qwenmax{} & 0.2440 & 0.8021 \\
Reasoning & \qwenmax{} & 0.2859 & 0.8150 \\
Conciseness & \qwenmax{} & 0.4019 & 0.9058 \\
Format & \qwenmax{} & 0.5020 & 0.9258 \\
\bottomrule
\end{tabular}%
\end{table}

The prompt-optimization diagnostic treats a finite set of candidate prompts as the search space and maximizes validation accuracy on a binary sentiment-classification task. Table~\ref{tab:prompt-llm-expert-diagnostics} shows that real prompt experts vary substantially in prior error. Several \deepseek{} roles have lower mean absolute error than the \qwenmax{} roles, while some high-confidence \qwenmax{} roles have high error. This supports the paper's main design choice: confidence should be treated as evidence to be calibrated, not as a probability of correctness.

\section{Candidate-Pool Scaling Diagnostic}
\label{sec:pool-size-scaling}

This diagnostic tests whether the synthetic stress-test conclusions depend on using a very small finite candidate pool. We use the two-objective synthetic benchmark with the mixed-objective expert scenario, vary the candidate pool size from \(100\) to \(10{,}000\), and keep the initial design, evaluation budget, and seed count fixed at \(6\), \(24\), and \(3\), respectively. Because the finite-pool oracle hypervolume changes slightly with pool size, Table~\ref{tab:pool-size-scaling} reports final hypervolume normalized by the full-pool oracle hypervolume for each seed.

\begin{table}[!htbp]
\centering
\small
\caption{Synthetic candidate-pool scaling diagnostic. Values are normalized final hypervolume \((\uparrow)\), reported as mean \(\pm\) SEM over three seeds.}
\label{tab:pool-size-scaling}
\resizebox{\linewidth}{!}{%
\begin{tabular}{@{}lllllll@{}}
\toprule
Pool & \method{} \((\uparrow)\) & \method{}, no conf. \((\uparrow)\) & qLogNEHVI \((\uparrow)\) & qLogEHVI \((\uparrow)\) & Fixed prior \((\uparrow)\) & Random search \((\uparrow)\) \\
\midrule
100 & 0.999 \(\pm\) 0.001 & 0.999 \(\pm\) 0.001 & 0.999 \(\pm\) 0.001 & \textbf{1.000 \(\pm\) 0.000} & 0.998 \(\pm\) 0.001 & 0.727 \(\pm\) 0.072 \\
300 & 0.997 \(\pm\) 0.001 & 0.997 \(\pm\) 0.001 & 0.987 \(\pm\) 0.012 & \textbf{0.998 \(\pm\) 0.001} & 0.806 \(\pm\) 0.174 & 0.702 \(\pm\) 0.112 \\
1000 & 0.983 \(\pm\) 0.006 & 0.982 \(\pm\) 0.005 & \textbf{0.996 \(\pm\) 0.001} & 0.833 \(\pm\) 0.164 & 0.727 \(\pm\) 0.155 & 0.781 \(\pm\) 0.080 \\
3000 & 0.983 \(\pm\) 0.002 & 0.984 \(\pm\) 0.003 & \textbf{0.987 \(\pm\) 0.002} & 0.985 \(\pm\) 0.007 & 0.973 \(\pm\) 0.006 & 0.649 \(\pm\) 0.028 \\
10000 & 0.967 \(\pm\) 0.012 & 0.971 \(\pm\) 0.009 & \textbf{0.991 \(\pm\) 0.002} & 0.989 \(\pm\) 0.004 & 0.976 \(\pm\) 0.005 & 0.762 \(\pm\) 0.032 \\
\bottomrule
\end{tabular}%
}
\end{table}

The calibrated standalone prior method remains in the same high-performance band as qLogEHVI and qLogNEHVI as the pool grows. It is close to the best method for pool sizes \(100\) and \(300\), remains above \(0.98\) normalized hypervolume through pool size \(3000\), and stays above random at \(10{,}000\). The largest-pool result also clarifies why the prior layer should be acquisition-agnostic: qLogNEHVI can be stronger than the standalone scalarized acquisition, motivating the qLogNEHVI plugin result in Table~\ref{tab:qlognehvi-prior-layer-scaling}.

\FloatBarrier

\section{Derivations and Algorithmic Details}
\label{sec:algorithm-details}

\subsection{Residual-Prior Bayesian Optimization}

Let \(p_{j,t}(x)\) be the current LLM-derived prior for objective \(j\). We model
\[
  f_j(x)=p_{j,t}(x)+r_j(x),
\]
where \(r_j\sim \mathcal{GP}(0,k_j)\). Given observations \(\obsset_t=\{(x_\tau,\mathbf{y}_\tau)\}_{\tau=1}^{t}\), the residual targets are
\[
  \tilde{y}_{j,\tau}=y_{j,\tau}-p_{j,t}(x_\tau).
\]
The GP posterior is fit to \(\{(x_\tau,\tilde{y}_{j,\tau})\}\), yielding residual posterior mean \(m^r_{j,t}(x)\) and standard deviation \(s^r_{j,t}(x)\). Therefore the predictive mean for the original objective is
\[
  \mathbb{E}[f_j(x)\mid \obsset_t]=p_{j,t}(x)+m^r_{j,t}(x).
\]
This decomposition lets the observed data check the LLM prior: if the prior is wrong, the residual GP learns systematic corrections; if the prior is harmful, the counterfactual prior gate can reduce or drop it.

\subsection{Reputation-Market Update}

For expert \(e\) and objective \(j\), define the scaled absolute error
\[
  \epsilon_{ej,t}=\frac{|\mu_{ej}(x_t)-y_{j,t}|}{s_{j,t}},
\]
where \(s_{j,t}\) is a running objective scale. The reward
\[
  R_{ej,t}=\clip(0.5-0.5\epsilon_{ej,t}^2,-2.0,0.5)
\]
is a shifted and clipped Gaussian log-score term. The shift allows accurate predictions to earn positive capital, while clipping prevents a single bad observation from dominating the market. Capital evolves as
\[
  K_{ej,t+1}=(1-\lambda)K_{ej,t}+\eta\,\tilde{c}_{ej}(x_t)R_{ej,t}.
\]
The expert weights are obtained by a temperature-controlled softmax over capital:
\[
  \alpha_{ej,t} =
  \frac{\exp(K_{ej,t}/T)}
       {\sum_{e'}\exp(K_{e'j,t}/T)}.
\]
This is equivalent to a discounted exponential-weights update in capital space, with objective-specific experts competing within each objective.

\subsection{Counterfactual Prior Gate}

The prior gate compares three possible prior actions for each objective:
\[
  a\in\{\mathrm{no\_conf},\mathrm{conf},\mathrm{drop}\}.
\]
For each action, the historical residual sequence is
\[
  r^{(a)}_{j,\tau}=y_{j,\tau}-p^{(a)}_{j,t}(x_\tau).
\]
The evidence score is the per-observation GP marginal log-likelihood
\[
  S_{j,t}(a)=\frac{1}{t}\log p(r^{(a)}_{j,1:t}\mid \obsx_t).
\]
The no-confidence action is used as the reference arm:
\[
  \ell_{j,t}(a)=\eta_p\left[S_{j,t}(a)-S_{j,t}(\mathrm{no\_conf})-\Delta_a\right].
\]
With \(n_t\) observations and count-scale \(\kappa_0\), the finite-data shrinkage factor is
\[
  \kappa_t=\sqrt{\frac{n_t}{n_t+\kappa_0}}.
\]
The prior-gate policy is
\[
  \pi^p_{j,t}=(1-\kappa_t)\mathbf{e}_{\mathrm{no\_conf}}
  +\kappa_t\,\softmax_a(\ell_{j,t}(a)).
\]
Here \(\mathbf{e}_{\mathrm{no\_conf}}\) denotes the unit vector for the no-confidence arm.
Thus early rounds default toward the conservative no-confidence prior, while later rounds allow the GP evidence to select confidence weighting or prior abstention.

\subsection{Counterfactual Update Gate}

The update gate decides whether confidence should scale future reputation rewards. The key observation is that the reputation market is deterministic once the reward history and confidence rule are fixed. We therefore maintain two parallel shadow markets at the beginning of round \(t\): \(g=0\) replays past expert rewards with confidence disabled, and \(g=1\) replays the same rewards with reported confidence enabled.

Let \(K^{(g)}_{ej,t}\) denote the capital of expert \(e\) for objective \(j\) in shadow market \(g\), after replaying observations up to round \(t-1\). The corresponding shadow expert weights are
\[
  \alpha^{(g)}_{ej,t}
  =
  \frac{\exp(K^{(g)}_{ej,t}/T)}
       {\sum_{e'}\exp(K^{(g)}_{e'j,t}/T)}.
\]
Using these weights, each shadow market forms the same no-confidence/confidence/drop prior mixture as the main prior gate:
\[
  p^{(g)}_{j,t}(x)
  =
  \pi^p_{j,t}(\mathrm{no\_conf})p^{(g,\mathrm{no\_conf})}_{j,t}(x)
  +
  \pi^p_{j,t}(\mathrm{conf})p^{(g,\mathrm{conf})}_{j,t}(x),
\]
where the drop-prior arm contributes zero. The nonzero component priors are
\[
  p^{(g,\mathrm{no\_conf})}_{j,t}(x)
  =
  \frac{\sum_e \alpha^{(g)}_{ej,t}\mu_{ej}(x)}
       {\sum_e \alpha^{(g)}_{ej,t}},
  \qquad
  p^{(g,\mathrm{conf})}_{j,t}(x)
  =
  \frac{\sum_e \alpha^{(g)}_{ej,t}c_{ej}(x)\mu_{ej}(x)}
       {\sum_e \alpha^{(g)}_{ej,t}c_{ej}(x)}.
\]
After observing \(\mathbf{y}_t\), the evaluated point \(x_t\) provides full-information feedback for both shadow markets:
\[
  L^u_{j,t}(g)=\frac{|y_{j,t}-p^{(g)}_{j,t}(x_t)|}{s_{j,t}}.
\]
This loss asks a narrow counterfactual question: if previous reputation updates had followed rule \(g\), would the resulting LLM prior have predicted the newly observed value more accurately?

The update-gate distribution follows Hedge:
\[
  \bar{L}^u_{j,t}(g)=L^u_{j,t}(g)-\frac{1}{2}\sum_{g'\in\{0,1\}}L^u_{j,t}(g'),
\]
\[
  \pi^u_{j,t+1}(g)=
  \frac{\pi^u_{j,t}(g)\exp(-\eta_u \bar{L}^u_{j,t}(g))}
       {\sum_{g'\in\{0,1\}}\pi^u_{j,t}(g')\exp(-\eta_u \bar{L}^u_{j,t}(g'))}.
\]
Centering the losses only subtracts an arm-independent constant within each objective and therefore does not change the softmax probabilities; it is used for numerical stability. The updated distribution \(\pi^u_{j,t+1}\) is used in future rounds.

For the real market update at round \(t\), we deliberately use the pre-observation gate probability
\[
  \rho_{j,t}=\pi^u_{j,t}(1),
\]
so that \(y_{j,t}\) is not used to decide how much confidence should scale its own reward. The resulting effective confidence multiplier is
\[
  \tilde{c}^{\,u}_{ej,t}
  =
  1+\rho_{j,t}\big(c_{ej}(x_t)-1\big),
\]
and the real capital update is
\[
  K_{ej,t+1}
  =
  \clip\!\left(
  (1-\lambda)K_{ej,t}
  +\eta\,\tilde{c}^{\,u}_{ej,t}R_{ej,t},
  K_{\min},K_{\max}
  \right).
\]
Here \(K_{\min}\) and \(K_{\max}\) are fixed clipping bounds that prevent a single update from making expert capital numerically dominant.

The same observation is then replayed into both shadow markets. Define the shadow confidence multiplier
\[
  \omega^{(0)}_{ej,t}=1,\qquad \omega^{(1)}_{ej,t}=c_{ej}(x_t).
\]
With the reputation reward \(R_{ej,t}\) from the observed prediction error, the two shadow markets evolve as
\[
  K^{(g)}_{ej,t+1}
  =
  \clip\!\left(
  (1-\lambda)K^{(g)}_{ej,t}
  +\eta\,\omega^{(g)}_{ej,t}R_{ej,t},
  K_{\min},K_{\max}
  \right),
  \qquad g\in\{0,1\}.
\]
Thus the newly observed point affects the real market through the old gate \(\rho_{j,t}\), while it affects the next gate only through the shadow-market losses and replay updates. Unlike a bandit meta-gate, this procedure observes the counterfactual loss of both update arms at every step and learns from an endogenous replay signal rather than from delayed, high-variance downstream hypervolume changes.

\FloatBarrier

\section{Pseudocode}

\begin{figure}[H]
\centering
\fbox{
\begin{minipage}{0.91\linewidth}
\small
\textbf{Algorithm 1: \method{} Evidence-Gated LLM Prior Layer}

\textbf{Input:} candidate pool \(\mathcal{X}\), cached expert priors \(\mu_{ej}(x)\), confidences \(c_{ej}(x)\), initial design size \(n_0\), budget \(B\).

\textbf{Initialize:} evaluate \(n_0\) candidates to form \(\obsset_{n_0}\) and \(\obsx_{n_0}\); initialize expert-objective capital \(K_{ej}\), prior-gate policy \(\pi^p_j\), update-gate policy \(\pi^u_j\), and shadow markets \(K^{(0)},K^{(1)}\).

\begin{enumerate}
  \item For \(t=n_0,\ldots,B-1\), compute \(\epsilon_{ej,t}\) and \(R_{ej,t}\) from the newest observation.
  \item Update the actual reputation market using the pre-observation update gate \(\rho_{j,t}=\pi^u_{j,t}(1)\).
  \item Score both update-gate shadow markets at \(x_t\), update \(\pi^u_j\) by full-information Hedge, and replay the observation into both shadows.
  \item For each objective \(j\), score the no-confidence, confidence, and drop-prior residual histories by fixed-kernel GP marginal likelihood; update \(\pi^p_j\).
  \item Aggregate \(p_{j,t}(x)\) with the three-arm prior mixture and fit residual GP surrogates to \(y_{j,\tau}-p_{j,t}(x_\tau)\).
  \item Select \(x_{t+1}=\arg\max_{x\in\mathcal{X}\setminus \obsx_t} A_t(x)\), evaluate it, and append \(x_{t+1}\) and \((x_{t+1},\mathbf{y}_{t+1})\) to form \(\obsx_{t+1}\) and \(\obsset_{t+1}\).
\end{enumerate}

\textbf{Output:} evaluated candidates and the final non-dominated set.
\end{minipage}
}
\caption{Pseudocode for the fixed \method{} prior layer used in the main experiments.}
\label{fig:appendix-pseudocode}
\end{figure}

\FloatBarrier

\section{Hyperparameter Sensitivity}

The main method intentionally fixes a compact hyperparameter set. We nevertheless include sensitivity diagnostics for the two most important counterfactual-gate parameters: the prior-abstention margin \(\Delta_{\mathrm{drop}}\) and the prior-gate inverse temperature \(\eta_p\).

\begin{table}[tbp]
\centering
\small
\caption{Drop-margin sensitivity on ESOL across LLM backends. Values are final hypervolume \((\uparrow)\); missing cells indicate runs not collected.}
\label{tab:appendix-drop-margin-sensitivity}
\resizebox{\linewidth}{!}{%
\begin{tabular}{@{}llll@{}}
\toprule
Backend & \(\Delta_{\mathrm{drop}}=0.05\) \((\uparrow)\) & \(\Delta_{\mathrm{drop}}=0.25\) \((\uparrow)\) & \(\Delta_{\mathrm{drop}}=1.0\) \((\uparrow)\) \\
\midrule
\qwenflash{} & \textbf{0.7561} & 0.7480 & 0.7515 \\
\qwenmax{} & 0.7470 & 0.7445 & \textbf{0.7507} \\
\deepseek{} & \textbf{0.7515} & N/A & 0.7500 \\
\bottomrule
\end{tabular}%
}
\end{table}

\begin{figure}[H]
  \centering
  \includegraphics[width=0.72\linewidth]{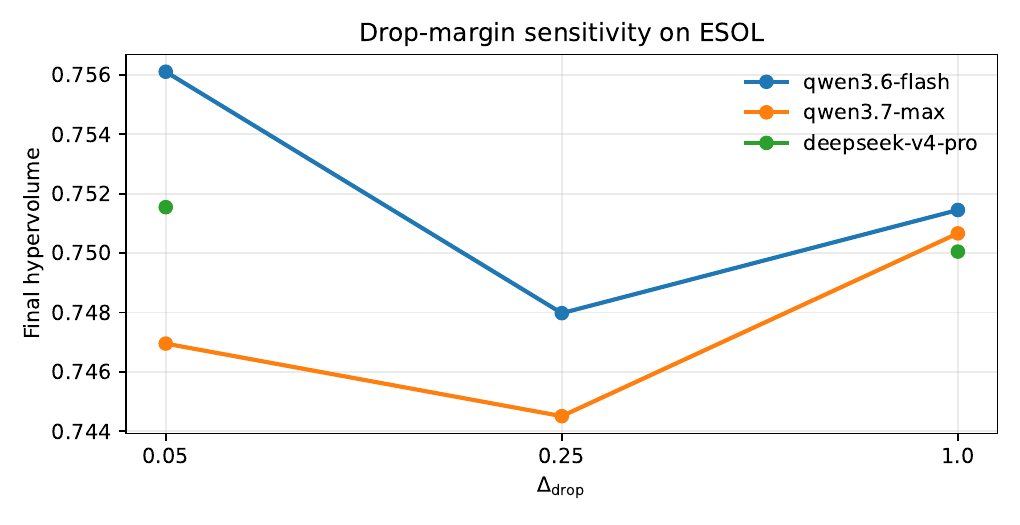}
  \caption{Drop-margin sensitivity on ESOL across three LLM backends. Bars show collected runs; missing backend--margin combinations are labeled N/A. The best margin differs across backends, which motivates our decision to keep a fixed, transparent margin in the main method and treat automatic margin selection as future work.}
  \label{fig:appendix-drop-margin-sensitivity}
\end{figure}

\begin{table}[tbp]
\centering
\small
\caption{Prior-gate temperature sensitivity. Values are final hypervolume \((\uparrow)\) for \(\eta_p=1\) and a sharper \(\eta_p=8\) gate.}
\label{tab:appendix-prior-eta-sensitivity}
\resizebox{\linewidth}{!}{%
\begin{tabular}{@{}lll@{}}
\toprule
Setting & \(\eta_p=1\) \((\uparrow)\) & \(\eta_p=8\) \((\uparrow)\) \\
\midrule
ESOL / \qwenflash{} & \textbf{0.7561} & 0.7416 \\
Synthetic / overconfident bad & 0.4788 & \textbf{0.4791} \\
Synthetic / correlated bad experts & \textbf{0.4461} & \textbf{0.4461} \\
\bottomrule
\end{tabular}%
}
\end{table}

The sensitivity results support two conclusions. First, the drop margin interacts with the LLM backend: \qwenflash{} prefers a more aggressive abstention margin in the collected runs, while \qwenmax{} and \deepseek{} are less aligned with that choice. Second, sharpening the prior-gate temperature does not consistently improve performance. This explains why we avoid treating the margin portfolio as the main method: one-step prior fit is not sufficient to choose the best exploration behavior.

\FloatBarrier

\section{Continuous-Space Diagnostic}
\label{sec:continuous-space-diagnostic}

The main experiments focus on finite candidate pools, where every cached LLM prior score is attached to a known candidate. To check whether the prior layer is inherently tied to this setting, we also ran a small continuous-space diagnostic on the two-objective Branin--Currin benchmark implemented in BoTorch \citep{balandat2020botorch}. This diagnostic uses all-good synthetic expert priors, not additional LLM calls. Candidates are selected from a continuous domain rather than from a fixed pool, and the expert-prior module is evaluated as a callable prior function rather than as a precomputed table.

\begin{table}[tbp]
\centering
\small
\caption{Continuous Branin--Currin diagnostic with all-good synthetic expert priors. Final hypervolume \((\uparrow)\) is reported as mean \(\pm\) SEM over three seeds; best sum \((\uparrow)\) is the mean final best objective-sum. The best value in each metric is bolded.}
\label{tab:continuous-branin-currin}
\resizebox{\linewidth}{!}{%
\begin{tabular}{@{}lll@{}}
\toprule
Method & Final \(\hv\) \((\uparrow)\) & Best sum \((\uparrow)\) \\
\midrule
Residual BO + \method{} & \textbf{0.9882 \(\pm\) 0.0005} & \textbf{1.9437} \\
qLogNEHVI & 0.9865 \(\pm\) 0.0027 & 1.9436 \\
qLogNEHVI + \method{} & 0.9709 \(\pm\) 0.0015 & 1.9178 \\
qLogNEHVI + fixed prior & 0.9707 \(\pm\) 0.0013 & 1.9178 \\
Vanilla MOBO & 0.8799 \(\pm\) 0.0953 & 1.7818 \\
\bottomrule
\end{tabular}%
}
\end{table}

\begin{figure}[H]
  \centering
  \includegraphics[width=0.78\linewidth]{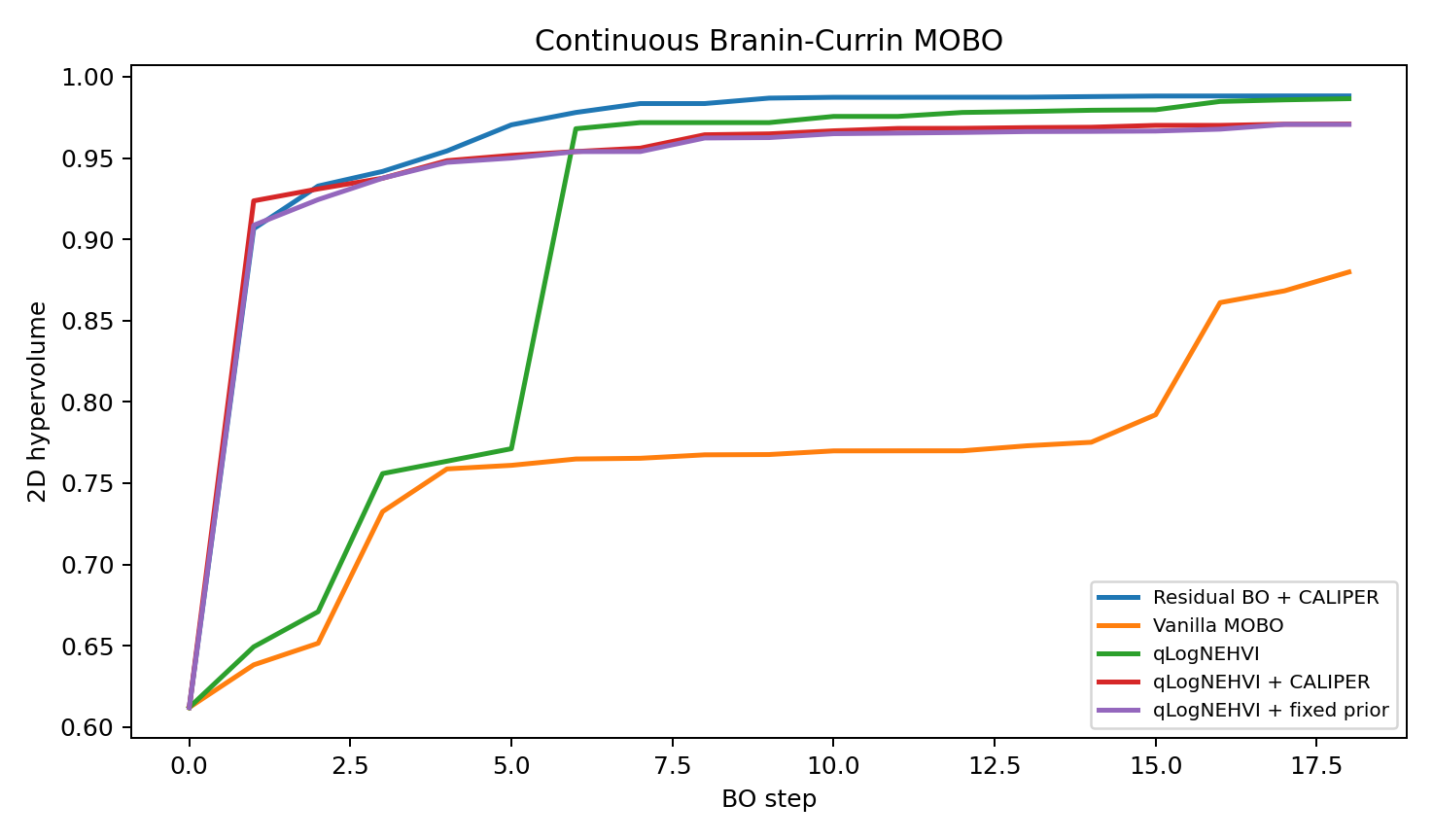}
  \caption{Continuous Branin--Currin hypervolume curves over three seeds with all-good synthetic expert priors. This diagnostic is included to test portability beyond finite candidate pools, not as a main performance claim.}
  \label{fig:continuous-branin-currin}
\end{figure}

Table~\ref{tab:continuous-branin-currin} and Figure~\ref{fig:continuous-branin-currin} show that the residual BO version of \method{} reaches final hypervolume comparable to qLogNEHVI on this continuous task. The qLogNEHVI prior-layer variants do not improve over qLogNEHVI here, which is consistent with the paper's broader caution: prior layers are useful when their prior signal and acquisition interface are well aligned, but the gate should not be interpreted as a guarantee of improvement on every acquisition and benchmark.

\end{document}